\documentclass[acmtog, nonacm]{acmart}

\setcitestyle{nosort,square}

\usepackage{mathrsfs}
\usepackage{tikz}
\usetikzlibrary{arrows.meta}
\usepackage{overpic}
\usepackage{pict2e}
\usepackage{dsfont}
\usepackage[most]{tcolorbox}
\usepackage[outline]{contour}
\usepackage{standalone}
\usepackage[normalem]{ulem}

\usepackage{microtype}      
\usepackage[capitalize]{cleveref}
\usepackage{xspace}         
\usepackage{mathtools}      
\usepackage{nicefrac,xfrac} 
\usepackage{mathtools}

\usepackage{stmaryrd}  
\usepackage{newfloat}  
\usepackage{savesym}
\usepackage{algorithm}
\usepackage{algorithmicx}
\usepackage[noend]{algpseudocode} 
\algrenewcommand\textproc{}
\usepackage{graphicx}
\usepackage{grffile}  
\usepackage{soul}
\usepackage{wrapfig}
\usepackage{bbm}
\usepackage{flushend}
\usepackage{tabularx}
\usepackage{ctable}            
\usepackage[a]{esvect}           
\usepackage{multirow}

\usepackage{wrapfig}
\usepackage{textcomp}    
\usepackage{enumitem}    

\definecolor{Red}{rgb}{1,0.45,0.45}
\definecolor{Green}{rgb}{0.44,0.76,0.33}
\definecolor{Blue}{rgb}{0.46,0.7,1}
\definecolor{DarkGreen}{rgb}{0.0,0.6,0.0}
\definecolor{DomainColor}{gray}{0}
\definecolor{ExtSpaceColor}{rgb}{0.86,0,0.86}
\definecolor{ExtSpaceColorTwo}{rgb}{0.86,0.5,0.86}
\definecolor{TargetSpaceColor}{hsb}{0.45,0.87,0.81}
\definecolor{TargetMeasureColor}{rgb}{0.3,0.3,0.3}
\definecolor{RedClassColor}{hsb}{0.99,0.62,0.9}
\definecolor{BlueClassColor}{hsb}{0.61,0.6,0.9}
\definecolor{PurpleClassColor}{rgb}{0.86,0.35,0.88}
\definecolor{ProjectionColor}{rgb}{0.95,0.57,0.00}

\newcommand{\Ignore}[1]{#1}  
\newcommand{\commentOutText}[1]{}

\newcommand{\revision}[1]{\Ignore{\textcolor[rgb]{0.0, 0.0, 0.0}{#1}}}

\newcommand{\mymath}[2]{\newcommand{#1}{\TextOrMath{$#2$\xspace}{#2}}}

\savesymbol{Comment}
\algrenewcommand\algorithmicindent{3mm}
\DeclareFloatingEnvironment[
    name=Alg.,
    placement=tbhp,
    within=none,
]{pseudocode}
\crefname{pseudocode}{Alg.}{Algs.}
\Crefname{pseudocode}{Algorithm}{Algorithms}

\newcommand{\AlgCommentTemplate}[2]{\hfill{\fontsize{9} {6}\selectfont\textcolor{DarkGreen}{\text{#1\;#2}}}}
\newcommand{\AlgComment}[1]{\AlgCommentTemplate{}{#1}}

\DeclareGraphicsExtensions{.ai,.pdf,.jpg,.png}

\DeclareDocumentCommand{\Outlined}{ O{black} O{white} O{0.55pt} m }{%
    \contourlength{#3}
    \contour{#2}{\textcolor{#1}{#4}}%
}

\newcommand{\undefinecolor}[1]{\expandafter\let\csname\string\color@#1\endcsname\undefined}

\newcommand{\mywfigurevspace}[4]{%

\begin{wrapfigure}{r}{#2\columnwidth}%
 \hspace{-0.7cm}%
 \vspace{#4}%
    \centering
    \includegraphics[width=#2\columnwidth]{#1}%
    \vspace{-.2cm}%
    \label{fig:#1}%
    \vspace{-0.3cm}%
\end{wrapfigure}%
}

\usepackage{bm}

\mymath{\Data}{\mathbf{x}}
\mymath{\DeltaTimeStep}{\Delta{t}}
\mymath{\alphat}{\alpha_{t}}
\mymath{\baralphat}{\bar{\alpha}_{t}}
\mymath{\alphatplusone}{\alpha_{t+1}}
\mymath{\alphatminusone}{\alpha_{t-1}}
\mymath{\NoisyDatat}{\mathbf{x}_{t}}
\mymath{\NoisyDatatplusone}{\mathbf{x}_{t+1}}
\mymath{\NoisyDatatminusone}{\mathbf{x}_{t-1}}
\mymath{\GaussianNoise}{\bm{\epsilon}}
\mymath{\GaussianBlueNoise}{\mathbf{b}}
\mymath{\NetworkOutputNoise}{\bm{\epsilon}_{\bm{\theta}}}
\mymath{\NetworkOutputScore}{\bm{s}_{\bm{\theta}}}
\mymath{\NetworkOutputIADB}{\bm{D}_{\bm{\theta}}}
\mymath{\NetworkOutput}{f_{\bm{\theta}}}
\mymath{\NetworkOutputFirst}{f_{\bm{\theta}}^{'}}
\mymath{\NetworkOutputSecond}{f_{\bm{\theta}}^{''}}

\mymath{\LossIADB}{\mathcal{L}_{IADB}}
\mymath{\LossOurs}{\mathcal{L}_{Ours}}
\mymath{\LossDDPM}{\mathcal{L}_{DDPM}}
\mymath{\SigmaMatrix}{\Sigma}
\mymath{\LMatrix}{L}

\mymath{\TotalTimesteps}{T}
\newcommand{\GaussianBlueNoiseText}{Gaussian blue noise\xspace}
\newcommand{\SupplementalDoc}{Supplemental document\xspace}
\newcommand{\SupplementalHTML}{Supplemental HTML\xspace}

\newcommand{\estimate}[1]{\hat{#1}}

\mymath{\density}d
\mymath{\pcf}g
\mymath{\pcfEstimate}{\estimate\pcf}
\mymath{\pcfTarget}{\bar\pcf}
\mymath{\pointPattern}P
\mymath{\kernel}{\kappa}\mymath{\densithKernel}{\kernel_\mathrm d}
\mymath{\correlationKernel}{\kernel_\mathrm g}
\mymath{\point}{\mathbf x}
\mymath{\location}{\mathbf y}
\mymath{\jitterlocation}{\mathbf q}
\mymath{\locationSpaced}{\location_{\hphantom{i}}}
\mymath{\guidance}{\mathcal H}
\mymath{\spatial}{\mathcal S}
\mymath{\pcfBandwidth}{\sigma}
\mymath{\editedPointPattern}{{\bar{\pointPattern}}}
\mymath{\pointCount}n
\mymath{\pcfSampleCount}m
\mymath{\radius}r
\mymath{\pcfRmax}{r_\mathrm{max}}
\mymath{\normal}{\mathcal N}
\mymath{\bandwidth}{\sigma}
\mymath{\intensity}{I}
\mymath{\inverseIntensity}{s}
\mymath{\densityScaling}{\alpha}
\mymath{\guide}h
\mymath{\guideBandwidth}{\Sigma}
\mymath{\densityMap}D
\mymath{\pcfMap}G
\mymath{\latentMap}Z
\mymath{\featureMap}F
\mymath{\editedFeatureMap}{\bar{\featureMap}}
\mymath{\edit}{\mathtt{edit}}
\mymath{\editedDensityMap}{\bar\densityMap}
\mymath{\editedPCFMap}{\bar\pcfMap}
\mymath{\editedLatentMap}{\bar\latentMap}
\mymath{\synthesisCost}{c_\mathrm{Syn}}
\mymath{\synthesisCostEstimate}{\estimate{c}_\mathrm{Syn}}
\mymath{\visualFeatureStats}{\mathbf v}
\mymath{\targetVisualFeatureStats}{\bar{\mathbf v}}
\mymath{\perceptualDistance}{\mathcal D}
\mymath{\latentCoordinate}{\mathbf z}
\mymath{\networkLossPixel}{\mathcal L_1}
\mymath{\networkLossGAN}{\mathcal L_\mathrm{adv}}
\mymath{\networkLossTotal}{\mathcal L_\mathrm{tot}}
\mymath{\locality}{\phi}
\mymath{\embeddingCost}{c_\mathrm{Emb}}
\mymath{\encoding}{f}
\mymath{\decoding}{f^{-1}}
\mymath{\pixelResolutionWidth}{256}
\mymath{\pixelResolutionHeight}{256}
\mymath{\learnignRate}{\lambda}

\mymath{\PowerSpectrum}{p}
\mymath{\PowerSpectrumBin}{b}
\mymath{\PowerSpectrumScale}{w}
\mymath{\PowerSpectrumGamma}{\gamma}
\mymath{\NumGMM}{n_\mathrm{GMM}}
\mymath{\mean}{\mu}
\mymath{\stddev}{\sigma}

\mymath{\cielab}{CIELAB}
\mymath{\lspace}{L}
\mymath{\aspace}{A}
\mymath{\bspace}{B}
\mymath{\abspace}{AB}

\mymath{\argmin}{\operatorname{arg\,min}}

\copyrightyear{2024}
\acmYear{2024}
\setcopyright{rightsretained}
\acmConference[SIGGRAPH Conference Papers '24]{Special Interest Group on Computer Graphics and Interactive Techniques Conference Conference Papers '24}{July 27-August 1, 2024}{Denver, CO, USA}
\acmBooktitle{Special Interest Group on Computer Graphics and Interactive Techniques Conference Conference Papers '24 (SIGGRAPH Conference Papers '24), July 27-August 1, 2024, Denver, CO, USA}
\acmDOI{10.1145/3641519.3657435}
\acmISBN{979-8-4007-0525-0/24/07}

\begin{document}


\title{Blue noise for diffusion models}


\author{Xingchang Huang}
\affiliation{%
  \institution{MPI Informatics, VIA Center}
  \city{Saarbr{\"u}cken}
  \country{Germany}
}
\email{xhuang@mpi-inf.mpg.de}

\author{Corentin Sala\"un}
\affiliation{%
  \institution{MPI Informatics}
  \city{Saarbr{\"u}cken}
  \country{Germany}
}
\email{csalaun@mpi-inf.mpg.de}

\author{Cristina Vasconcelos}
\affiliation{%
  \institution{Google DeepMind}
  \city{London}
  \country{UK}
}
\email{crisnv@google.com}

\author{Christian Theobalt}
\affiliation{%
  \institution{MPI Informatics, VIA Center}
  \city{Saarbr{\"u}cken}
  \country{Germany}
}  
\email{theobalt@mpi-inf.mpg.de}

\author{Cengiz {\"O}ztireli}
\affiliation{%
  \institution{Google Research, University of Cambridge}
  \city{Cambridge}
  \country{UK}
}
\email{cengizo@google.com}

\author{Gurprit Singh}
\affiliation{%
  \institution{MPI Informatics, VIA Center}
  \city{Saarbr{\"u}cken}
  \country{Germany}
}  
\email{gsingh@mpi-inf.mpg.de}

\renewcommand{\shortauthors}{X. Huang, C. Sala\"un, C. Vasconcelos, C. Theobalt, C. {\"O}ztireli, G. Singh}


\begin{teaserfigure}

\newcommand{\PlotSingleImage}[1]{%
        \begin{scope}
            \clip (0,0) -- (2.5,0) -- (2.5,2.5) -- (0,2.5) -- cycle;
            \path[fill overzoom image=figures/#1] (0,0) rectangle (2.5cm,2.5cm);
        \end{scope}
        \draw (0,0) -- (2.5,0) -- (2.5,2.5) -- (0,2.5) -- cycle;
        
}

\newcommand{\PlotSingleImageWithLine}[1]{%
        \begin{scope}
            \clip (0,0) -- (2.5,0) -- (2.5,2.5) -- (0,2.5) -- cycle;
            \path[fill overzoom image=figures/#1] (0,0) rectangle (2.5cm,2.5cm);
        \end{scope}
        \draw (0,0) -- (2.5,0) -- (2.5,2.5) -- (0,2.5) -- cycle;
        \draw[dashed] (0, -0.13) -- (2.5, -0.13);
}

\newcommand{\TwoColumnFigure}[2]{%
    \begin{tabular}{c@{\;}c@{}}
        \hspace*{-2.5mm}
        \begin{tikzpicture}[scale=0.563]
            \PlotSingleImage{#1}
        \end{tikzpicture}
         & 
         \begin{tikzpicture}[scale=0.563]
            \PlotSingleImage{#2}
        \end{tikzpicture}
    \end{tabular}%
}
\newcommand\scalevalue{0.72}    
\small
\hspace*{-4mm}
\begin{tabular}{c@{\;}c@{}}
\begin{tabular}{c@{\;}c@{\;}c@{\;}c@{\;}c@{\;}c@{\;}c@{\;}c@{}}
\rotatebox{90}{\hspace{0.1cm} \scriptsize \citet{heitz2023iterative}}
&
\begin{tikzpicture}[scale=\scalevalue]
\PlotSingleImage{cat_res128_iadb/gwn_img02000_step0.png}
\end{tikzpicture}
&
\begin{tikzpicture}[scale=\scalevalue]
\PlotSingleImage{cat_res128_iadb/gwn_img02000_step50.png}
\end{tikzpicture}
&
\begin{tikzpicture}[scale=\scalevalue]
\PlotSingleImage{cat_res128_iadb/gwn_img02000_step125.png}
\end{tikzpicture}
&
\begin{tikzpicture}[scale=\scalevalue]
\PlotSingleImage{cat_res128_iadb/gwn_img02000_step175.png}
\end{tikzpicture}
&
\begin{tikzpicture}[scale=\scalevalue]
\PlotSingleImage{cat_res128_iadb/gwn_img02000_step200.png}
\end{tikzpicture}
&
\begin{tikzpicture}[scale=\scalevalue]
\PlotSingleImage{cat_res128_iadb/gwn_img02000_step225.png}
\end{tikzpicture}
&
\begin{tikzpicture}[scale=\scalevalue]
\PlotSingleImage{cat_res128_iadb/gwn_img02000_step250.png}
\end{tikzpicture}
\\[-0.4mm]
\rotatebox{90}{\hspace{0.1cm} \scriptsize Blue noise (Ours)}
&
\begin{tikzpicture}[scale=\scalevalue]
\PlotSingleImage{cat_res128_iadb/gbn_img00000_step0.png}
\end{tikzpicture}
&
\begin{tikzpicture}[scale=\scalevalue]
\PlotSingleImage{cat_res128_iadb/gbn_img00000_step50.png}
\end{tikzpicture}
&
\begin{tikzpicture}[scale=\scalevalue]
\PlotSingleImage{cat_res128_iadb/gbn_img00000_step125.png}
\end{tikzpicture}
&
\begin{tikzpicture}[scale=\scalevalue]
\PlotSingleImage{cat_res128_iadb/gbn_img00000_step175.png}
\end{tikzpicture}
&
\begin{tikzpicture}[scale=\scalevalue]
\PlotSingleImage{cat_res128_iadb/gbn_img00000_step200.png}
\end{tikzpicture}
&
\begin{tikzpicture}[scale=\scalevalue]
\PlotSingleImage{cat_res128_iadb/gbn_img00000_step225.png}
\end{tikzpicture}
&
\begin{tikzpicture}[scale=\scalevalue]
\PlotSingleImage{cat_res128_iadb/gbn_img00000_step250.png}
\end{tikzpicture}
\\[-0.4mm]
\rotatebox{90}{\hspace{0.0cm} \scriptsize Time-varying (Ours)}
&
\begin{tikzpicture}[scale=\scalevalue]
\PlotSingleImageWithLine{cat_res128_iadb_linear_teaser/gwn2gbn_img02000_step0.png}
\end{tikzpicture}
&
\begin{tikzpicture}[scale=\scalevalue]
\PlotSingleImageWithLine{cat_res128_iadb_linear_teaser//gwn2gbn_img02000_step50.png}
\end{tikzpicture}
&
\begin{tikzpicture}[scale=\scalevalue]
\PlotSingleImageWithLine{cat_res128_iadb_linear_teaser/gwn2gbn_img02000_step125.png}
\end{tikzpicture}
&
\begin{tikzpicture}[scale=\scalevalue]
\PlotSingleImageWithLine{cat_res128_iadb_linear_teaser/gwn2gbn_img02000_step175.png}
\end{tikzpicture}
&
\begin{tikzpicture}[scale=\scalevalue]
\PlotSingleImageWithLine{cat_res128_iadb_linear_teaser/gwn2gbn_img02000_step200.png}
\end{tikzpicture}
&
\begin{tikzpicture}[scale=\scalevalue]
\PlotSingleImageWithLine{cat_res128_iadb_linear_teaser/gwn2gbn_img02000_step225.png}
\end{tikzpicture}
&
\begin{tikzpicture}[scale=\scalevalue]
\PlotSingleImageWithLine{cat_res128_iadb_linear_teaser/gwn2gbn_img02000_step250.png}
\end{tikzpicture}
\\[-0.4mm]
\rotatebox{90}{\hspace{0.1cm} \scriptsize Our noise masks}
&
\begin{tikzpicture}[scale=\scalevalue]
\PlotSingleImage{interpolated_noises_res128/interpolated_noise_t250.png}
\end{tikzpicture}
&
\begin{tikzpicture}[scale=\scalevalue]
\PlotSingleImage{interpolated_noises_res128/interpolated_noise_t200.png}
\end{tikzpicture}
&
\begin{tikzpicture}[scale=\scalevalue]
\PlotSingleImage{interpolated_noises_res128/interpolated_noise_t125.png}
\end{tikzpicture}
&
\begin{tikzpicture}[scale=\scalevalue]
\PlotSingleImage{interpolated_noises_res128/interpolated_noise_t075.png}
\end{tikzpicture}
&
\begin{tikzpicture}[scale=\scalevalue]
\PlotSingleImage{interpolated_noises_res128/interpolated_noise_t050.png}
\end{tikzpicture}
&
\begin{tikzpicture}[scale=\scalevalue]
\PlotSingleImage{interpolated_noises_res128/interpolated_noise_t025.png}
\end{tikzpicture}
&
\begin{tikzpicture}[scale=\scalevalue]
\PlotSingleImage{interpolated_noises_res128/interpolated_noise_t000.png}
\end{tikzpicture}
\\[-0.4mm]
~ &
$t=250$ & 
$t=200$ &
$t=125$ &
$t=75$ &
$t=50$ &
$t=25$ &
$t=0$
\\[-0.4mm]
\end{tabular}
&
\newcommand\scalevalShortcut{0.729}
\vline width 1.0pt
\begin{tabular}{c@{\;}c@{}}
\begin{tikzpicture}[scale=\scalevalShortcut]
\PlotSingleImage{teaser/gwn_img18000_step250.png}
\end{tikzpicture}
&
\begin{tikzpicture}[scale=\scalevalShortcut]
\PlotSingleImage{teaser/gwn2gbn_img18000_step250.png}
\end{tikzpicture}
\\[-0.4mm]
\begin{tikzpicture}[scale=\scalevalShortcut]
\PlotSingleImage{teaser/gwn_img16000_step250.png}
\end{tikzpicture}
&
\begin{tikzpicture}[scale=\scalevalShortcut]
\PlotSingleImage{teaser/gwn2gbn_img16000_step250.png}
\end{tikzpicture}
\\[-0.4mm]
\begin{tikzpicture}[scale=\scalevalShortcut]
\PlotSingleImage{teaser/gwn_img05200_step250.png}
\end{tikzpicture}
&
\begin{tikzpicture}[scale=\scalevalShortcut]
\PlotSingleImage{teaser/gwn2gbn_img05200_step250.png}
\end{tikzpicture}
\\[-0.4mm]
\begin{tikzpicture}[scale=\scalevalShortcut]
\PlotSingleImage{teaser/gwn_img00000_step250.png}
\end{tikzpicture}
&
\begin{tikzpicture}[scale=\scalevalShortcut]
\PlotSingleImage{teaser/gwn2gbn_img00000_step250.png}
\end{tikzpicture}
\\[-0.4mm]
\footnotesize \citet{heitz2023iterative} &
\large \textbf{Ours}
\\[-0.4mm]
\end{tabular}
\end{tabular}

  \vspace{-.2cm}    
  \caption{
  In conventional diffusion-based generative modeling, data is corrupted by adding Gaussian (random) noise (as illustrated in the top row). Here, we explore the alternative approach of using correlated Gaussian noise in diffusion-based generative modeling. 
  Different correlation can be used such as blue noise (second row) or a time-varying correlations (third row). 
  Bottom row shows the corresponding noise mask for time-varying example at different time steps. 
  Rightmost two columns show visual comparisons on generated images (from the same initial noise) between~\citet{heitz2023iterative} (using only Gaussian noise) and Ours (using time-varying noise). Our generated images are more natural-looking and detailed with less artifacts.
}
  \label{fig:Teaser}
\end{teaserfigure}


\begin{abstract}

Most of the existing diffusion models use Gaussian noise for training and sampling across all time steps, which may not optimally account for the frequency contents reconstructed by the denoising network.
Despite the diverse applications of correlated noise in computer graphics, its potential for improving the training process has been underexplored.
In this paper, we introduce a novel and general class of diffusion models taking correlated noise within and across images into account. 
More specifically, we propose a time-varying noise model to incorporate correlated noise into the training process, as well as a method for fast generation of correlated noise mask.
Our model is built upon deterministic diffusion models and utilizes blue noise to help improve the generation quality compared to using Gaussian white (random) noise only. 
Further, our framework allows introducing correlation across images within a single mini-batch to improve gradient flow.
We perform both qualitative and quantitative evaluations on a variety of datasets using our method, achieving improvements on different tasks over existing deterministic diffusion models in terms of FID metric.
Code will be available at 
\textcolor[RGB]{30,144,255}{\url{https://github.com/xchhuang/bndm}}.

\end{abstract}



\maketitle


\section{Introduction}

Since the groundbreaking work by \citet{sohl2015deep, ho2020denoising, song2019generative}, there has been extensive research on diffusion models. These models have demonstrated superior performance in terms of generative quality and training stability compared to Generative Adversarial Networks (GANs) \cite{dhariwal2021diffusion} for image synthesis. Additionally, diffusion models can be trained to perform various tasks such as text-to-image synthesis, image inpainting, image super-resolution, and image editing. 

Typically, a diffusion model consists of two processes: forward and backward. In the forward process, the model gradually adds noise to an original data point (e.g., an image), transforming it into a random noise pattern. In the backward process, the model learns to reconstruct the original data from this noise using a denoising neural network. The denoising network initially focuses on reconstructing the coarse shape and structure (low-frequency components) in the early time steps. As the time steps decrease, the denoising network progressively refines the details (high-frequency components). This behavior indicates that diffusion models generate data in a coarse-to-fine manner and have a hidden relationship with frequency components. 

Despite these advancements, there is limited research on the relationship between this behavior and the noise used during the forward and backward processes. Most existing diffusion models rely solely on Gaussian noise, also known as uncorrelated Gaussian noise or Gaussian white noise, as its frequency power spectrum spans all frequencies (similar to the \emph{white} color). While correlated noise has not been thoroughly examined in diffusion models, there has been some relevant research that has delved into this domain. For example, \citet{rissanen2022generative} propose a diffusion process inspired by heat dissipation to explicitly control frequencies. Similarly, \citet{voleti2022score} suggest using non-isotropic noise as a replacement for isotropic Gaussian noise in score-based diffusion models. Despite their potential, both methods face limitations regarding the quality of the generated images, which could explain their limited adoption in mainstream models.

In this paper, we propose a new diffusion model that supports a diffusion process with time-varying noise. Our goal is to utilize correlated noise, such as blue noise (\cite{ulichney1987digital}), to enhance the generative process. Blue noise is characterized by a power spectrum with no energy in its low-frequency region. Our focus is on blue noise masks (\cite{ulichney1999void}), which provide noise profiles with blue noise properties. We propose using these blue noise \emph{masks} to design a time-varying noise for diffusion-based generative modeling. Generating such correlated noise masks for diffusion is a time-consuming process, as it may require generating thousands to millions of masks on the fly. To address this issue, we propose an efficient method to generate \GaussianBlueNoiseText masks on the fly for both low-dimensional and high-dimensional images. In summary, our contributions are as follows: 

\begin{itemize}[leftmargin=3.4mm]
\item We propose a framework that investigates the impact of correlated noise and correlation across training images on generative diffusion models.
\item Based on our framework, we introduce a deterministic diffusion process with time-varying noise, allowing control over the correlation introduced in the model at each step. 
\item We overcome the computational challenges of generating correlated noise masks by introducing a real-time mask generation approach. 
\item By interpolating Gaussian noise and \GaussianBlueNoiseText~using our proposed time-varying noise model, our model outperforms existing deterministic models like IADB~\cite{heitz2023iterative} or DDIM~\cite{song2020denoising} in various image generation tasks.
\end{itemize}

\section{Related Work}

\paragraph{Blue noise}
Blue noise is a type of noise characterized by high frequency signals and the absence of low frequencies. It has found numerous applications in computer graphics. One such application is the utilization of blue noise masks, originally introduced by~\citet{Ulichney1993VoidandclusterMF}, for image dithering to enhance their perceptual quality. Blue noise masks are also employed in Monte Carlo rendering to improve the distribution of error, as demonstrated by \citet{georgiev2016blue} and \citet{heitz2019distributing}. The relationship between blue noise and denoising in rendering has been further explored by  \citet{chizhov2022perceptual} and  \citet{Salaun:2022:ScalableMultiClassSampling}, revealing that combining blue noise with a low pass filter can reduce perceptual errors. To leverage the advantageous denoising properties of blue noise masks, we propose to use them as additive noise to corrupt the data for diffusion-based generative modeling.

\paragraph{Diffusion models}
There are various formulations for image generation by diffusion models, including stochastic~\cite{ho2020denoising, song2019generative,song2020score} and deterministic ones~\cite{song2020denoising,heitz2023iterative}.
Diffusion models also extend beyond image generation to video generation~\cite{ho2022video} and 3D content generation~\cite{poole2022dreamfusion}.
More comprehensive reviews can be found in the surveys by \citet{cao2024survey} and \citet{po2023state}.

Diffusion models are known to be slow to train, as well as slow in the generative process. 
How to speed up the generative process to generate images in a few steps becomes an increasingly important research topic\revision{~\cite{lu2022dpm,liu2022flow,liu2023instaflow, salimans2022progressive, karras2022elucidating, karras2023analyzing, song2023consistency, luo2023latent}}.
Orthogonal to reducing the number of inference steps, some work focus on developing a more general framework that can support various types of noise addition~\cite{jolicoeur2023diffusion} or image corruption operations~\cite{bansal2024cold}. 
Some work explicitly take frequency of the image content into account to model the generative process in a coarse-to-fine manner~\cite{rissanen2022generative, phung2023wavelet}.
However, there exists limited work researching the how the frequency of noise used in image corruption can make an impact on the denoising process in diffusion-based generative modeling.
To understand this problem, we propose a framework utilizing correlated noise to improve the denoising process.

\section{Our method}

\begin{figure}[t]
    \centering
\newcommand{\Plotdistrib}[1]{%
    \begin{scope}
            \clip (0,0) -- (2.5,0) -- (2.5,1.25) -- (0,1.25) -- cycle;
            \path[fill overzoom image=figures/motivation/#1] (0,0) rectangle (2.5,1.25);
        \end{scope}
        \draw (0,0) -- (2.5,0) -- (2.5,1.25) -- (0,1.25) -- cycle;
}
\newcommand{\PlotSingleImageBlue}[1]{%
    \begin{tikzpicture}[scale=0.5]    
        \begin{scope}
            \clip (0,0) -- (2.5,0) -- (2.5,2.5) -- (0,2.5) -- cycle;
            \path[fill overzoom image=figures/#1] (0,0) rectangle (2.5cm,2.5cm);
        \end{scope}
        \draw [blue] (0,0) -- (2.5,0) -- (2.5,2.5) -- (0,2.5) -- cycle;
    \end{tikzpicture}%
}
\newcommand{\PlotSingleImageRed}[1]{%
    \begin{tikzpicture}[scale=0.5]
        \begin{scope}
            \clip (0,0) -- (2.5,0) -- (2.5,2.5) -- (0,2.5) -- cycle;
            \path[fill overzoom image=figures/#1] (0,0) rectangle (2.5cm,2.5cm);
        \end{scope}
        \draw [red] (0,0) -- (2.5,0) -- (2.5,2.5) -- (0,2.5) -- cycle;
    \end{tikzpicture}%
}

\newcommand{\TwoColumnFigure}[2]{%
    \begin{tabular}{c@{\;}c@{}}
        \hspace*{-2.5mm}
        \begin{tikzpicture}[scale=0.535]    
            \PlotSingleImageBlue{#1}
        \end{tikzpicture}
         & 
         \begin{tikzpicture}[scale=0.535]
            \PlotSingleImageRed{#2}
        \end{tikzpicture}
    \end{tabular}%
}

\newcommand{\scaleval}{3.3} 

\small
\hspace*{-4mm}
\begin{tabular}{c@{\hskip 0.1in}c@{\hskip 0.1in}c@{\;}}
\multicolumn{3}{c}{\begin{tikzpicture}[scale=\scaleval]
    \Plotdistrib{motiv_diffusion_distribution.png}
    \Plotdistrib{motiv_diffusion_points.png}
    \Plotdistrib{gradient_flow_colored_lines.pdf}
    \begin{scope}
        \draw [blue,-{Stealth[width=2mm]}](0.835,0.32) -- (0.4,-0.1); 
        \draw [violet,-{Stealth[width=2mm]}](1.235,0.38) -- (1.26,-0.1);
        \draw [red,-{Stealth[width=2mm]}](1.67,0.37) -- (2.12,-0.1);
    \end{scope}
\end{tikzpicture}}
\\
\TwoColumnFigure{interpolated_noises_res128/interpolated_noise_t200.png}{motivation/motaivation_x1_deblended_200.png}
&
\TwoColumnFigure{interpolated_noises_res128/interpolated_noise_t125.png}{motivation/motaivation_x1_deblended_125.png}
&
\TwoColumnFigure{interpolated_noises_res128/interpolated_noise_t050.png}{motivation/motaivation_x1_deblended_50.png}
\\
$t=200$ & $t=125$ & $t=50$
\end{tabular}
    \vspace{-2mm} 
    \caption{
        Schema of diffusion process using our time-varying noise. The diffusion transforms initial noise distribution (blue) into the target data distribution (red). Five examples are shown with the intermediates diffusion steps between the two distributions. For one of the data we illustrate the intermediates time steps with the current expect result and noise. The evolution of the noise from random to blue noise is visible as well as the quality of the expected result.
    }
    \label{fig:motivation}
\end{figure}

A generative model based on diffusion comprises two key processes: a forward process and a backward process. In the forward process, noise, denoted as $\GaussianNoise$, is introduced to corrupt an initial image, $\Data_0$, by scaling it with a factor determined by a discrete-time parameter, $t$. Here, $\Data_0$ represents a real image sampled from the training data distribution, denoted as $p_0$. The time step, $t$, ranges from 0 to $T-1$, where $T$ is the total number of discrete time steps. The corrupted image, along with the corresponding time step $t$, is then used as input to train a neural network, $\NetworkOutput(\Data_t, t)$.

In the backward process, the trained network is employed to denoise pure noise and generate new images. \Cref{fig:motivation} illustrates this process. Starting from Gaussian noise (blue distribution), the image iteratively passes through the network to eventually yield a fully denoised image, aligning with the target distribution (red distribution). Intermediate steps of the process involve a mixture of noise and image. Three examples are visible in the figure. As more time steps are executed (with $t$ closer to 0), the image quality improves, and more details emerge. In this example, the intermediate noise transitions from Gaussian noise to Gaussian blue noise following a time-varying schedule following \cref{subsec:time-varying_diffusion}.\\

This section explores correlations across two different axes: across pixels in the noise and across images within a mini-batch.
To demonstrate the impact of correlations between noise masks and images, 
we build a deterministic diffusion process with time-varying noise following the work by~\citet{heitz2023iterative} namely, the IADB method. For the sake of simplicity and fair one-to-one comparison, our method was developed on top of IADB while preserving its characteristics and hyperparamenters other than the ones described as new on our method. But our method is general enough that could be potentially explored on top of other existing generative diffusion process.

For IADB, the forward and backward processes and the objective function are defined as the following:
\begin{align}
\label{eq:iadb}
    \NoisyDatat &= \alphat \GaussianNoise + (1 - \alphat) \Data_0
    \\
    \NoisyDatatminusone &= \NoisyDatat + (\alphat - \alphatminusone) \NetworkOutput(\Data_t, t)
    \\
    \LossIADB &= \sum_t (\NetworkOutput(\Data_t, t) - (\Data_0 - \GaussianNoise))^2
\end{align}
With $\Data_0$ a target image, $\GaussianNoise \sim \mathcal{N}(\bm{0},\,\boldsymbol{I})$ a random Gaussian noise and $\alphat$ and $\alphatminusone$ two blending coefficient. The network model is referred as $\NetworkOutput$ and take 2 input parameters : a corrupted image $\NoisyDatat$ and the timestep $t$.
A stochastic formulation of IADB also exist but this work will focus on the deterministic variant for its stability.

\subsection{Correlated noise}

In a deterministic diffusion process noise masks are used as initialization of the backward process to generate images and at each training step to corrupt target images.
Mask generation during training is a critical factor and must meet specific requirements. The process must be stochastic to produce different masks at each iteration, reducing overfitting and increasing diversity in the generated results. Mask generation also required to be of fast computation as it is employed at every training step.

Gaussian noise naturally meet this requirements but it is not the case for all correlated mask method. 
In particular, IADB uses mask generated from a multivariate Gaussian distribution with zero mean and identity covariance matrix. 
A method to create correlated noise such as blue noise requires a non-identity covariance matrix.
The covariance matrix of the blue noise mask can be estimated from a collection of masks. We employed simulated annealing using objective function from \citet{Ulichney1993VoidandclusterMF} to generate ten thousand blue noise masks. While this method yields high-quality masks, it does require significant optimization time.
Then the blue noise correlation matrix $\Sigma$ can be computed by averaging the respective correlation matrix of the example masks.

To create a noise mask with a specified covariance matrix $\Sigma$, the Cholesky decomposition is applied to $\Sigma$, resulting in a lower-triangular matrix $\LMatrix$ ($\LMatrix\LMatrix^T = \Sigma$). Finally, the random vector is multiplied with $\LMatrix$ to produce the desired noise mask efficiently:
\begin{align}
    \label{eq:GaussianBlueNoise}
    \GaussianBlueNoise &= \LMatrix\GaussianNoise
\end{align}
where $\GaussianNoise \sim \mathcal{N}(0, I)$ is a unit-variance Gaussian distribution.
%
\begin{figure}[t!]
    \centering

\newcommand{\PlotSingleImage}[1]{%
        \begin{scope}
            \clip (0,0) -- (2.5,0) -- (2.5,2.5) -- (0,2.5) -- cycle;
            \path[fill overzoom image=figures/noise_visualization/#1] (0,0) rectangle (2.5cm,2.5cm);
        \end{scope}
        \draw (0,0) -- (2.5,0) -- (2.5,2.5) -- (0,2.5) -- cycle;
}

\newcommand{\TwoColumnFigure}[2]{%
    \begin{tabular}{c@{\;}c@{}}
        \hspace*{-2.5mm}
        \begin{tikzpicture}[scale=0.563]
            \PlotSingleImage{#1}
        \end{tikzpicture}
         & 
         \begin{tikzpicture}[scale=0.563]
            \PlotSingleImage{#2}
        \end{tikzpicture}
    \end{tabular}%
}

\newcommand{\scaleval}{1.04}
\small
\hspace*{-3mm}
\begin{tabular}{c@{\;}c@{\;}c@{\;}c@{\;}c@{}}
%
\raisebox{-.45\height}{\begin{tikzpicture}[scale=\scaleval]
\PlotSingleImage{bn_res64.png}
\end{tikzpicture}}
&
\centering \huge =
&
\raisebox{-.45\height}{\begin{tikzpicture}[scale=\scaleval]
\PlotSingleImage{cov_mat_L_res16.png}
\end{tikzpicture}}
&
\centering $\times$
&
\raisebox{-.45\height}{\begin{tikzpicture}[scale=\scaleval]
\PlotSingleImage{wn_res64.png}
\end{tikzpicture}}
\\
Gaussian blue noise $\mathbf{b}$ &
~&
Triangular Matrix $\LMatrix$ &
~&
Gaussian noise $\bm{\epsilon}$
\\ [-0.4mm]
\end{tabular}

    \vspace{-2mm} 
    \caption{
        To generate on-the-fly Gaussian blue noise masks $\GaussianBlueNoise$ (leftmost), we pre-compute a lower triangular matrix $\LMatrix$. 
        We then multiply this matrix with Gaussian noise $\GaussianNoise$ to obtain $\GaussianBlueNoise=\LMatrix \GaussianNoise$.
        For Gaussian noise of size $64\times64$,
        the lower triangular matrix has a size of $64^2 \times 64^2$. 
        Here we show a $64 \times 64$ zoom-in version of the matrix to better visualize the positive and negative correlations shown as the white and black lines, respectively. 
    }
    \label{fig:gwn_gbn}
\end{figure}
%
\Cref{fig:gwn_gbn} shows one realization of Gaussian blue noise mask generated using~\cref{eq:GaussianBlueNoise}.
Each row or column in the \LMatrix~represent a pixel index of the noise mask. 
Each cell in the \LMatrix~represents the correlation strength between the pixels in the noise mask. 
Positive values correspond to bright cells that represent positive correlations. 
Similarly, negative values correspond to dark cells representing negative correlations.
For each pixel, only its adjacent pixels have values far from zero while other non-adjacent pixels have values close to zero, as indicated by the white and black lines.
Note that we use the name Gaussian blue noise for $\GaussianBlueNoise$ in this paper, but is different from the Gaussian Blue Noise method from~\citet{ahmed2022gaussian}.

\paragraph{Higher-dimension} 
Matrix-vector multiplication are computationally expensive when the \LMatrix matrix is high-dimensional. Increasing directly the matrix size to generate higher-dimensional noise becomes slower than generating (uncorrelated) Gaussian noise using a modern machine learning framework such as PyTorch~\cite{paszke2017automatic}.
As noise generation is used at every training step, the overhead should remain minimal. 
Therefore, for generating higher-dimensional noise masks, we adapt \citet{kollig2002efficient} method for producing our blue noise masks in which a padding of a set of lower dimensional masks is applied.

More specifically, to generate a batch of high-dimensional Gaussian blue noise mask, a larger batch is generated at resolution $64^2$ using~\cref{eq:GaussianBlueNoise} with $\LMatrix \in \mathbb{R}^{64^2\times64^2}$. 
Then the $64^2$ masks are padded into larger tiles to get Gaussian blue noise masks at higher dimensions.
Thus, the computation overhead for generating a $128^2$ resolution Gaussian blue noise is negligible \revision{($\sim0.0002$ seconds)}. \Cref{fig:Teaser} shows examples of Gaussian blue noise with resolution $128^2$ generated by padding. Using padding for higher-dimensional mask creates seams between the padded tiles. This artifact is practically not visible and is compensated by the low overhead of the method.
We provide the masks at different resolutions, with corresponding frequency power spectra for Gaussian blue noise in the \SupplementalDoc Sec. 3, demonstrating that the property of blue noise is preserved at different resolutions using our padding method.

\subsection{Diffusion model with time-varying noise}
\label{subsec:time-varying_diffusion}

\begin{figure}[t]
    \centering

\newcommand{\PlotSingleImage}[1]{%
        \begin{scope}
            \clip (0,0) -- (2.5,0) -- (2.5,2.5) -- (0,2.5) -- cycle;
            \path[fill overzoom image=figures/interpolated_noises/#1] (0,0) rectangle (2.5cm,2.5cm);
        \end{scope}
        \draw (0,0) -- (2.5,0) -- (2.5,2.5) -- (0,2.5) -- cycle;
}

\newcommand{\TwoColumnFigure}[2]{%
    \begin{tabular}{c@{\;}c@{}}
        \hspace*{-2.5mm}
        \begin{tikzpicture}[scale=0.563]
            \PlotSingleImage{#1}
        \end{tikzpicture}
         & 
         \begin{tikzpicture}[scale=0.563]
            \PlotSingleImage{#2}
        \end{tikzpicture}
    \end{tabular}%
}

\newcommand\scalevalue{0.645}    

\small
\hspace*{-4mm}
\begin{tabular}{c@{\;}c@{\;}c@{\;}c@{\;}c@{\;}c@{\;}c@{\;}c@{\;}c@{\;}c@{\;}c@{}}
~ &
$t=999$ & 
$t=800$ &
$t=600$ &
$t=300$ &
$t=0$
\\
\rotatebox{90}{\hspace{0.5cm} \scriptsize Noises}
&
\begin{tikzpicture}[scale=\scalevalue]
\PlotSingleImage{interpolated_noise_t999.png}
\end{tikzpicture}
&
\begin{tikzpicture}[scale=\scalevalue]
\PlotSingleImage{interpolated_noise_t800.png}
\end{tikzpicture}
&
\begin{tikzpicture}[scale=\scalevalue]
\PlotSingleImage{interpolated_noise_t600.png}
\end{tikzpicture}
&
\begin{tikzpicture}[scale=\scalevalue]
\PlotSingleImage{interpolated_noise_t300.png}
\end{tikzpicture}
&
\begin{tikzpicture}[scale=\scalevalue]
\PlotSingleImage{interpolated_noise_t000}
\end{tikzpicture}
\\[-0.4mm]
\rotatebox{90}{\hspace{0.1cm} \scriptsize Power spectra}
&
\begin{tikzpicture}[scale=\scalevalue]
\PlotSingleImage{ps_interpolated_noise_t900.png}
\end{tikzpicture}
&
\begin{tikzpicture}[scale=\scalevalue]
\PlotSingleImage{ps_interpolated_noise_t800.png}
\end{tikzpicture}
&
\begin{tikzpicture}[scale=\scalevalue]
\PlotSingleImage{ps_interpolated_noise_t600.png}
\end{tikzpicture}
&
\begin{tikzpicture}[scale=\scalevalue]
\PlotSingleImage{ps_interpolated_noise_t300.png}
\end{tikzpicture}
&
\begin{tikzpicture}[scale=\scalevalue]
\PlotSingleImage{ps_interpolated_noise_t000.png}
\end{tikzpicture}
\\[-0.4mm]
\end{tabular} 
    \vspace{-2mm} 
    \caption{
        Visualization of linearly interpolated noises from Gaussian noise to Gaussian blue noise at resolution $64^2$ (top row), and the corresponding frequency power spectra (bottom row).
    }
    \label{fig:interpolated_noises}
\end{figure}

With a single matrix $\LMatrix$ only a single correlation can be generated.
For a diffusion model, controlling the amount of correlation introduced within the model at each time step is necessary. 
A time-varying \LMatrix can be compute from 2 fixed matrices encoding 2 correlation types:
\begin{align}
    \label{eq:time-varying_L}
    \LMatrix_t &= \gamma_t \LMatrix_w + (1 - \gamma_t) \LMatrix_b,
\end{align}
where $\LMatrix_w$ and $\LMatrix_b$ represent 2 different matrices and $\gamma_t$ the blending coefficient. Based on it, the forward process is defined as:
\begin{align}
    \label{eq:forward}
    \Data_t &= \alphat (\LMatrix_t \GaussianNoise) + (1 - \alphat) \Data_0
\end{align}
With this forward process, Gaussian noise and Gaussian blue noise are interpolated based on the time step $t$.
More generally, this model supports smooth interpolation any two type of noises based on $\gamma_t$.
\Cref{fig:interpolated_noises} shows an example of linear interpolating from Gaussian noise to Gaussian blue noise. 
The corresponding frequency power spectra, computed by Discrete Fourier Transform, show that the energy in the low-frequency region is decreasing from left to right.

\begin{algorithm}[t]
\small     
\caption{\normalsize Pseudocode for \texttt{forward} method}
\begin{algorithmic}[1]
\label{alg:forward}
    \Function{\texttt{forward}}{$\LMatrix_w,\LMatrix_b, \gamma_t$}\AlgComment{\footnotesize $\leftarrow$ White and blue noise matrices $\LMatrix$,}
    \State $\GaussianBlueNoise \gets \texttt{get\_noise}(\LMatrix_w,\LMatrix_b, \gamma_t)$\AlgComment{\footnotesize  blending coefficient $\gamma_t$}
    \State $\alphat \sim \mathcal{U}(0,1)$
    \State $\Data_{0} \sim p_0$ 
    \State $\Data_t \gets \alphat\GaussianBlueNoise + (1-\alphat) \Data_{0}$ \AlgComment{\footnotesize $\leftarrow$ \cref{eq:forward}}
    \State \Return $\Data_t$
\EndFunction
\end{algorithmic}
\end{algorithm}
\begin{algorithm}[t]
\small     
\caption{\normalsize Pseudocode for \texttt{backward} method}
\begin{algorithmic}[1]
\label{alg:backward}
    \Function{\texttt{backward}}{~} 
    \State $\Data \sim \mathcal{N}(\bm{0},\,\boldsymbol{I})$
    \For{$t \gets T$ to $1$} 
        \State $\alphat \gets \texttt{get\_alpha}(t)$\hspace*{6.5em}%
        \rlap{\raisebox{-1.0ex}{\smash{$\left.\begin{array}{@{}c@{}}{}\\{}\end{array}\color{DarkGreen}\right\}%
          \color{DarkGreen}
          \begin{tabular}{l}
          \footnotesize User defined\\$\alpha$-scheduler    
          \end{tabular}$}}}
        \State $\alphatminusone \gets \texttt{get\_alpha}(t-1)$
        \State $\gamma_t \gets \texttt{get\_gamma}(t)$\hspace*{9em}%
        \rlap{\raisebox{-1.0ex}{\smash{$\left.\begin{array}{@{}c@{}}{}\\{}\end{array}\color{DarkGreen}\right\}%
          \color{DarkGreen}
          \begin{tabular}{l}\footnotesize\cref{eq:get_alpha}
          \end{tabular}$}}}
        \State $\gamma_{t-1} \gets \texttt{get\_gamma}(t-1)$
        \State $\Data \gets \Data + (\alphat - \alphatminusone)\NetworkOutputFirst(\Data, t)$\\
        \quad \quad \quad \quad $+ (\gamma_t - \gamma_{t-1})\NetworkOutputSecond(\Data, t)$ \AlgComment{\footnotesize $\leftarrow$ \cref{eq:ours_backward}}
    \EndFor
    \State \Return $\Data$
\EndFunction
\end{algorithmic}
\end{algorithm}
\begin{algorithm}[t]
\small     
\caption{\normalsize Pseudocode for time-varying \texttt{get\_noise} method}
\begin{algorithmic}[1]
\label{alg:noise}
\Function{\texttt{get\_noise}}{$\LMatrix_w,\LMatrix_b, \gamma_t$} 
    \State $\GaussianNoise \sim \mathcal{N}(\bm{0},\,\boldsymbol{I})$
    \State $\LMatrix_t \gets \gamma_t\LMatrix_w + (1-\gamma_t)\LMatrix_b$ \AlgComment{\footnotesize $\leftarrow$ \cref{eq:time-varying_L}}
    \State $\GaussianBlueNoise \gets \LMatrix_t \GaussianNoise$ \AlgComment{\footnotesize $\leftarrow$ \cref{eq:GaussianBlueNoise}}
    \State \Return $\GaussianBlueNoise$ 
\EndFunction
\end{algorithmic}
\end{algorithm}

Next, the forward process need to be inverted to define the backward process.
Based on the definitions of the \LMatrix~and the forward process, we can derive the backward step as the following:
\begin{align}
\label{eq:ours_backward}
    \Data_{t-1} &= \Data_t + (\alphat - \alphatminusone) (\Data_0 - \LMatrix_t \GaussianNoise) + (\gamma_t - \gamma_{t-1}) \alphatminusone (\LMatrix_b \GaussianNoise - \LMatrix_w \GaussianNoise)
\end{align}
Detailed derivations can be found in the \SupplementalDoc Sec. 1.
Here, $\LMatrix_w$ is an identity matrix representing Gaussian (white) noise, $\LMatrix_b$ is the matrix defined in~\cref{eq:GaussianBlueNoise}. 
When $\LMatrix_b = \LMatrix_w$, our model falls back to IADB. When $\LMatrix_b \neq \LMatrix_w$, we obtain a more general model with time-varying noises.

In IADB, the network is designed to learn only the term $\Data_0 - \LMatrix_t \GaussianNoise$, where $\LMatrix_t$ is simply an identity matrix in their case.
Here we can train the network to learn both terms in~\cref{eq:ours_backward}.
A brute force way to achieve this would be using two neural networks. However, this is not practical as it will introduce significantly more computation than IADB.
We choose to output a 6-channel image, representing the two terms as two 3-channel images, noted as \NetworkOutputFirst($\Data_t$, t) and \NetworkOutputSecond($\Data_t$, t), respectively.
The desired network output becomes $\Data_0 - \LMatrix_t \GaussianNoise$ and $\alphatminusone (\LMatrix_b \GaussianNoise - \LMatrix_w \GaussianNoise)$.
Therefore, the loss function becomes:
\begin{flalign}
\label{eq:ourloss}
    \LossOurs = \sum_t &((\NetworkOutputFirst(\Data_t, t) - (\Data_0 - \LMatrix_t\GaussianNoise))^2  
    \nonumber
    \\  
    &+ \frac{\gamma_{t}-\gamma_{t-1}}{\alphat - \alphatminusone}(\NetworkOutputSecond(\Data_t, t) - \alphatminusone(\LMatrix_b \GaussianNoise - \LMatrix_w \GaussianNoise))^2))
\end{flalign}

Note that though our model is trained with time-varying noises, it is still deterministic during the backward process.
The backward process starts with an initial Gaussian noise and no additional noises is required in the intermediate time steps. 
Instead, the network learns to guide the backward process in a time-varying denoising manner.

The procedures of forward, backward and noise generation are summarized in~\cref{alg:noise,alg:forward,alg:backward}.
In~\cref{alg:backward}, we consider \texttt{get\_alpha} ($\alpha$-scheduler) as a linear function ($\alphat=t/T$) following~\citet{heitz2023iterative}, but it can be non-linear functions as well.
Next, we define \texttt{get\_gamma} as a general sigmoid-based function in~\cref{eq:get_alpha}.
The weighted term $(\gamma_{t} - \gamma_{t-1}
) / (\alphat - \alphatminusone)$ in~\cref{eq:ourloss} automatically accounts for the difference between $\alpha$-schedulers and $\gamma$-scheduler. 
When $\gamma_{t} - \gamma_{t-1}$ is small, the contribution of \NetworkOutputSecond($\Data_t$, t) decreases. This is consistent to the backward process described in~\cref{eq:ours_backward}, where \NetworkOutputSecond($\Data_t$, t) is less important when $\gamma_{t} - \gamma_{t-1}$ is small.

\paragraph{Noise scheduler.}

Inspired by the study from~\citet{chen2023importance}, the scheduler choice have an importance impact in particular with increasing image resolution. 
We parameterize \texttt{get\_gamma}, the $\gamma$-scheduler, as a sigmoid-based function to control the interpolation between two noises.
More specifically, the $\gamma$-scheduler is parameterized by 3 parameters: $start, end, \tau$ according to~\citet{chen2023importance}:
\begin{align}
    \label{eq:get_alpha}
    sigmoid\left(\frac{start + (end - start)*t/T}{\tau}\right)
\end{align}
where $sigmoid(x) = 1 / (1 + e^{-x})$ and $t$ is the time step.

Since it is not known how to set $start$, $end$ and $\tau$ in advance, we consider optimizing them in addition to the network parameters where $start \in [-3, 0)$, $end \in (0, 3]$, $\tau \in [0.01, 1000.0]$.
During the initial experiments, we found that $start$ and $end$ converged stably to around 0 and 3, while $\tau$ converged to around 0.2 or kept increasing, depending on the image resolution.
Meanwhile, we found that optimizing these 3 parameters took extra epochs to converge and made the training of the network more difficult, due to their changes over epochs.
\mywfigurevspace{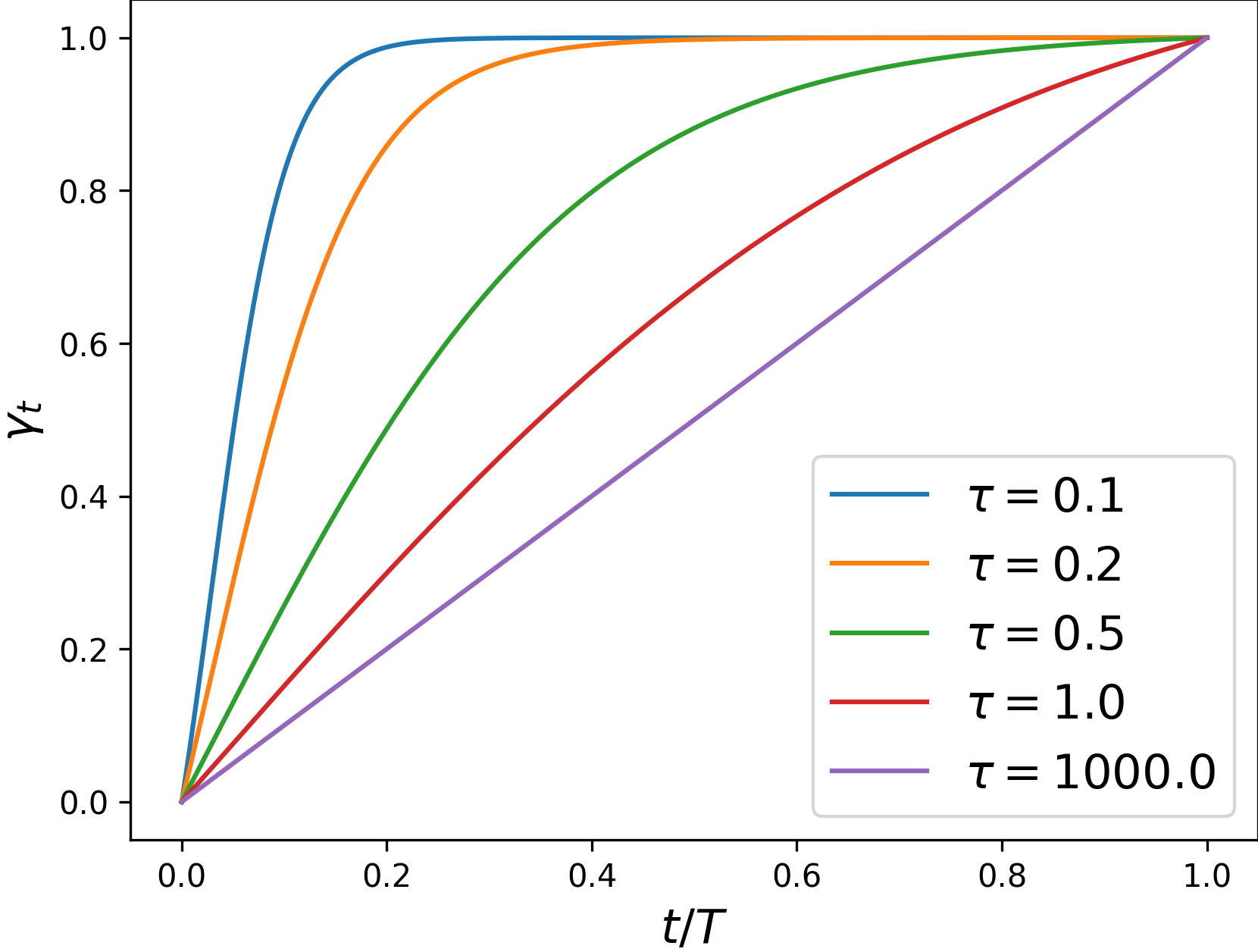}{0.5}{}{-0.0cm}
To make the choice of the 3 parameters more practical, we choose to fix $start=0, end=3$ and set $\tau$ based on the image resolution: $\tau=0.2$ for $128^2$, $\tau=1000$ for $64^2$ images.
The curves of the $\gamma$-scheduler with different $\tau$ values are shown in the inset image.
We summarize the values of the 3 parameters we use in the \SupplementalDoc Sec. 2 for all experiments.

\paragraph{Discussion}

\begin{figure}[h!]
    \centering

\newcommand{\PlotSingleImage}[1]{%
        \begin{scope}
            \clip (0,0) -- (2.5,0) -- (2.5,2.5) -- (0,2.5) -- cycle;
            \path[fill overzoom image=figures/3-axis_visualization/#1] (0,0) rectangle (2.5cm,2.5cm);
        \end{scope}
        \draw (0,0) -- (2.5,0) -- (2.5,2.5) -- (0,2.5) -- cycle;
}
\newcommand{\PlotSingleImageBlue}[1]{%
        \begin{scope}
            \clip (0,0) -- (2.5,0) -- (2.5,2.5) -- (0,2.5) -- cycle;
            \path[fill overzoom image=figures/#1] (0,0) rectangle (2.5cm,2.5cm);
        \end{scope}
        \draw [blue] (0,0) -- (2.5,0) -- (2.5,2.5) -- (0,2.5) -- cycle;
}
\newcommand{\PlotSingleImageRed}[1]{%
        \begin{scope}
            \clip (0,0) -- (2.5,0) -- (2.5,2.5) -- (0,2.5) -- cycle;
            \path[fill overzoom image=figures/3-axis_visualization/#1] (0,0) rectangle (2.5cm,2.5cm);
        \end{scope}
        \draw [red] (0,0) -- (2.5,0) -- (2.5,2.5) -- (0,2.5) -- cycle;
}

\newcommand{\TwoColumnFigure}[2]{%
    \begin{tabular}{c@{}}
        \hspace*{-2.5mm}
        \begin{tikzpicture}[scale=0.535]    
            \PlotSingleImageBlue{#1}
        \end{tikzpicture}
        \\ 
        \begin{tikzpicture}[scale=0.535]
            \PlotSingleImageRed{#2}
        \end{tikzpicture}
    \end{tabular}%
}

\newcommand{\scaleval}{1.3} 
\small
\hspace*{-4mm}
\begin{tabular}{c@{\;}c@{}c@{}}

Uncorrelated mapping 
&
\hspace{-2mm} \vspace{-1mm} Rectified mapping
&
~
\\[1mm]
\multirow{2}{*}[0.52in]{
\begin{tikzpicture}[scale=\scaleval]
    \PlotSingleImage{diffusion_distribution.png}
    \PlotSingleImage{diffusion_points.png}
\end{tikzpicture}
}
&
\multirow{2}{*}[0.52in]{
\begin{tikzpicture}[scale=\scaleval]
    \PlotSingleImage{diffusion_distribution.png}
    \PlotSingleImage{diffusion_points_remap.png}
    \begin{scope}
        \draw [blue,-{Stealth[width=2mm]},dashed](1.7,2.1) -- (2.65,2.1); 
        \draw [red,-{Stealth[width=2mm]},dashed](2.015,0.26) -- (2.65,0.26);
    \end{scope}
\end{tikzpicture}
}
&
\begin{tikzpicture}[scale=\scaleval/2.08]
\PlotSingleImageBlue{noise_visualization/bn_res128.png}
\end{tikzpicture}
\\
&
&
\begin{tikzpicture}[scale=\scaleval/2.08]
\PlotSingleImageRed{0000169.png}
\end{tikzpicture}

\\
(a) & \vspace{-1mm} (b) & (c)
\end{tabular}
    \vspace{-2mm} 
    \caption{
        Visualization of the impact of rectified mapping on mini-batch noise-mapping pairing. Blue and red points respectively represent the randomly selected noise and target image sampled in a given mini-batch from the underlying blue and red distribution. Standard practice (a) consist in a random mapping between the noise and target images. Our rectified mapping (b) improved it by reducing distance between the data pair. One example of noise and target from the mini-batch is visible (c). This example have been generated using our noise mask and mapping algorithm.
    }
    \label{fig:3-axis_visualization}
\end{figure}

Our time-varying noise model provides more capacity to choose a data-dependent scheduler for $\gamma_{t}$ to improve the denoising process.
One potential issue is that we need extra epochs to search for optimal parameters for the $\gamma$-scheduler.
To alleviate this problem, we propose a practical solution, which is to fix $\tau$ based on our initial optimization.
But how to choose the $\gamma$-scheduler in a more efficient way requires more study in the future.

\subsection{Data sample correlation using rectified mapping}

The previous paragraphs have demonstrated the use of correlated noise across pixels to enhance the diffusion process. Correlation can also be employed within a single mini-batch to improve the mapping between noise and the target image. 

Inspired by Rectified flow~\cite{liu2022flow} and Instaflow~\cite{liu2023instaflow}, correlation can be utilized to rectify the paired noise-image. 
\revision{~\Cref{fig:3-axis_visualization} visualizes a single mini-batch of paired data sample $\Data_0$ (red distribution) and noise $\GaussianBlueNoise$ (blue distribution) during a training iteration, and our rectified mapping.}
Previous work (\cref{fig:3-axis_visualization}(a)) applies a random mapping between $\Data_0$ and $\GaussianBlueNoise$. The noise-data mapping can be improved by applying an in-context stratification before feeding them into the forward process (\cref{fig:3-axis_visualization}(b)). This rectified mapping reduces the distance between each noise and its target image, resulting in a more direct trajectory. 
\revision{To find the mapping, we compute the squared distance between noises and images at the individual pixel level using the L2 norm.}
Then, for each $\GaussianBlueNoise$, the $\Data_0$ with the shortest distance that has not yet been used is selected. 
\revision{This improved mapping ensures that a specific image will consistently be associated with the same type of noise during the training process, resulting in a smooth gradient flow across time steps.
}

\section{Experiments}
%
\subsection{Implementation details}
\revision{We use CelebA~\cite{CelebAMask-HQ}, AFHQ-Cat~\cite{choi2020stargan} and LSUN-Bedroom~\cite{yu15lsun} datasets, with different resolutions, for unconditional/conditional image generation. More details on experimental setup can be found in Supplemental document Sec. 2.}

Our framework is implemented in Pytorch~\cite{paszke2017automatic} based on the official implementations from~\citet{song2020denoising,heitz2023iterative}.
We use 2D U-Net~\cite{ronneberger2015u} implemented in diffusers library~\cite{von-platen-etal-2022-diffusers}. More details about the network architecture and training details, including the values of $\tau$ in~\cref{eq:get_alpha}, can be found in the \SupplementalDoc Sec. 2.
Regarding the hyperparameters in diffusion models, we use \TotalTimesteps= 1000 for training and \TotalTimesteps= 250 testing.
To optimize the network parameters, we use AdamW optimizer~\cite{loshchilov2017decoupled} with learning rate 0.0001.
We use 4 NVIDIA Quadro RTX 8000 (48 GB) GPUs to train and test on all datasets.

For evaluation, we use FID~\cite{heusel2017gans}, Precision and Recall~\cite{kynkaanniemi2019improved} to measure the generative quality of all models.
The metrics are computed using the implementation from~\cite{stein2023exposing}, with Inception-v3 network~\cite{szegedy2016rethinking} as backbone. 
We generate 30k images to compute FID, Precision and Recall for all datasets.

\subsection{Image generation}

We compare our method with two existing deterministic diffusion models DDPM~\citet{ho2020denoising}, DDIM~\cite{song2020denoising} and IADB~\cite{heitz2023iterative} on unconditional image generation.
To ensure equitable comparisons, we employ identical initial Gaussian noise across all methods throughout the generative process.
Note that DDIM is trained using the diffusers library~\cite{von-platen-etal-2022-diffusers} with the same training setup compared to IADB and Ours.
\revision{
We also compare with stochastic diffusion models DDPM~\cite{ho2020denoising} and IHDM~\cite{rissanen2022generative} for completeness using the same number of time steps. Our method shows consistent improvement over the two methods.
}

The results on AFHQ-Cat ($64^2$), LSUN-Church ($64^2$), and Celeba ($64^2$) at the same resolution ($64^2$) are shown in Figure~\ref{fig:church_res64_sequence_gwn_gbn}. Our method exhibits the blue noise effect starting from approximately $t=75$, which visually distinguishes it from other methods. In terms of the generated images at time step $t=0$, our method produces images with less distortion around the pillars of the building and more detailed content around the windows and doors. In addition to visual comparisons, the quantitative evaluations presented in Table~\ref{tab:quantitative} demonstrate consistent improvements of our method over IADB and DDIM for datasets with a resolution of $64^2$.

\begin{table}[t]
\centering
\caption{
Quantitative FID score comparisons among \revision{IHDM~\cite{rissanen2022generative}, DDPM~\cite{ho2020denoising},} DDIM~\cite{song2020denoising}, IADB~\cite{heitz2023iterative}, and our method across diverse datasets. Notably, our approach exhibits improvements over IADB on every evaluated dataset. While our method is outperformed by DDIM on only one dataset, it's worth noting that IADB also performs poorly on the same dataset. Additional metrics are provided in the \SupplementalDoc Sec. 3.
}
\resizebox{8.5cm}{!}{
\begin{tabular}{c|>{\centering\arraybackslash}m{0.06\textwidth}>{\centering\arraybackslash}m{0.06\textwidth}>{\centering\arraybackslash}m{0.06\textwidth}>
{\centering\arraybackslash}m{0.06\textwidth}>{\centering\arraybackslash}m{0.06\textwidth}}
\toprule
FID ($\downarrow$) & IHDM & DDPM & DDIM & IADB & Ours\\
\hline
AFHQ-Cat ($64^2$) & 11.02 & 9.75 & 9.82 & 9.19 & \textbf{7.95} \\
AFHQ-Cat ($128^2$) & 15.40 & 12.41 & 10.73 & 10.81 & \textbf{9.47} \\
CelebA ($64^2$)  & 14.30 & 8.56 & 9.26 & 7.53 & \textbf{7.05}\\
CelebA ($128^2$)  & 36.93 & 15.06 & \textbf{11.92} & 20.71 & 16.38\\
LSUN-Church ($64^2$) & 17.76 & 13.07 & 16.46 & 13.12 & \textbf{10.16} \\
\bottomrule
\end{tabular}
}
\label{tab:quantitative}
\end{table}

For higher-resolution results, we observe the difference in the generated content starting at approximately $t=75$, as depicted in ~\cref{fig:celeba_res128_sequence_gwn_gbn}. Towards the end of the backward process, around $t=25$, we begin to notice the emergence of the blue noise effect as we use $\tau=0.2$ (\cref{eq:get_alpha}). To examine the details more closely, please zoom in. Additionally, we offer a \SupplementalHTML viewer where the intermediate generated images can be interactively visualized at various time steps during the backward process. In terms of realism, our generated images exhibit improved quality in regions such as hair, mouth, and eyes compared to IADB, as shown in ~\cref{fig:celeba_res128_sequence_gwn_gbn}. When compared to DDIM, our method achieves similar visual quality. Quantitative results on the CelebA ($128^2$) dataset, as presented in \cref{tab:quantitative}, demonstrate that DDIM outperforms IADB and our method. This outcome is attributed to DDIM employing a different expression for $\alphat$ in \cref{eq:iadb}, as discussed by in \citet{heitz2023iterative}. Depending on the dataset, DDIM may outperform IADB due to the distinct choice of $\alphat$. However, this is not a limitation for either IADB or our method. 

Additionally, our framework can also consider rectified mapping across images during training.~\cref{tab:Correlated_batch_mapping} provides the FID score with and without data correlation with respect to the number of diffusion steps, tested on AFHQ-Cat ($64^2$). Rectified mapping in the mini-batch achieves lower FID when the number of steps remains low, but a slightly higher FID when increasing the step count.

We provide additional results in the \SupplementalDoc Sec. 3, including \revision{detailed timing of the Gaussian blue noise generation and the backward process, and} a nearest neighbors test to confirm that our method does not overfit the training data.
\revision{\paragraph{Extension to other diffusion models}
Our method can be extended to DDIM, supported by derivations and preliminary results in Supplemental document Sec. 1, 3.
Also, our method can be incorporated into LDM~\cite{rombach2022high} for high-res image generation. As shown in~\cref{fig:cat_res512_compare_gwn_gbn}.
ours generates more realistic eyes and has better FID ($\textbf{11.45}<12.19$) compared with IADB on AFHQ-Cat ($512^2$).} 
Nevertheless, developing a new framework based on other models, necessitates additional effort, which we defer to future work.

\begin{table}[h]
\centering
\caption{
Comparing the impact of rectified mapping during training on AFHQ-Cat ($64^2$). FID scores ($\downarrow$) are provided with and without rectified mapping across different step counts. Correlation in the mini-batch results in lower FID at low steps but higher during slow diffusion.
}
\label{tab:Correlated_batch_mapping}
\resizebox{\columnwidth}{!}{%
\begin{tabular}{ccccccc}
\toprule
Diffusion step count (t) & 1 & 2 & 4 & 16 & 128 & 250 \\
\midrule
Uncorrelated batch mapping & 402.4 & 330.5 & 130.8 & 14.3 & \textbf{7.9} & \textbf{7.95}\\
Correlated batch mapping & \textbf{397.0} & \textbf{321.8} & \textbf{118.3} & \textbf{12.4} & 8.0 & 8.2 \\
\bottomrule
\end{tabular}
}
\end{table}
\subsection{Conditional image generation}
Besides unconditional generation from noise, our model also works for conditional image generation, such as image super-resolution, by simply concatenating the conditional low-resolution image with the noisy image as input. 

\Cref{fig:celeba_res128_superres} shows comparisons between IADB and Ours for image super-resolution on the LSUN-Church dataset from resolution $32^2$ to $128^2$. Our method outperforms IADB quantitatively according to SSIM~\cite{wang2004image}, PSNR and mean squared error (MSE). Our method outperforms IADB in terms of fidelity to the reference, as evidenced by its lower MSE. Visually, IADB tends to introduce excessive details, particularly in the bottom portion of the first image. Our method also effectively preserves straight lines throughout the image.
The quantitative results of all image-super resolution experiments can be found in the \SupplementalDoc Sec. 3, showing that our method consistently outperforms IADB.

\subsection{Ablation study and analysis}

\paragraph{Combinations of noises.} 
To confirm that Gaussian blue noise works due to its high-frequency property, we replace Gaussian blue noise by Gaussian red noise, a low-frequency noise visualized in~\cref{fig:gwn_gbw_grn}.
Red noises are generated using the same method~\cite{Ulichney1993VoidandclusterMF} by simply maximizing the objective function instead of minimizing it. 
Then we compute the corresponding covariance matrix and lower triangular matrix so that we can generate Gaussian red noise for our framework.
As shown in~\cref{fig:cat_res128_compare_gwn_gbn_grn}, using Gaussian red noise in our framework failed to recover the fine details due to its low-frequency property.
\Cref{tab:ablationNoises} shows that replacing \GaussianBlueNoiseText~by Gaussian red noise dramatically drops the Precision, while the Recall is comparable.
This is consistent to the visual observations in~\cref{fig:cat_res128_compare_gwn_gbn_grn} as Precision mainly measures the realism of the generated images.

\begin{table}[h]
\centering
\caption{
Ablation study on different combinations of noises using our framework on AFHQ-Cat ($128^2$). 
The last two rows mean blending Gaussian noise with Gaussian red or blue noise using the $\gamma$-scheduler with $\tau=0.2$.
}
\vspace{-2.5mm}
\label{tab:ablationNoises}
\resizebox{\columnwidth}{!}{%
\begin{tabular}{cccc}
\toprule
Noise model & FID ($\downarrow$) & Precision ($\uparrow$) & Recall ($\uparrow$) \\
\midrule
Ours (Gaussian white noise only) & 10.81 & \textbf{0.78} & 0.31 \\
Ours (\GaussianBlueNoiseText~only) & 17.61 & 0.59 & 0.18 \\ 
Ours (Gaussian white+red noise) & 13.64 & 0.67 & \textbf{0.34} \\ 
Ours (Gaussian white+blue noise) & \textbf{9.47} & \textbf{0.78} & \textbf{0.34} \\
\bottomrule
\end{tabular}
}
\end{table}

Another option is to use only \GaussianBlueNoiseText, which has been shown in~\cref{fig:Teaser} (second row). 
The final generated images are less realistic compared to IADB and Ours. 
The visual quality is also consistent with the quantitative metrics as shown in~\cref{tab:ablationNoises}.
But it is worth mentioning that in the case of using only \GaussianBlueNoiseText, we can observe that the content of the image appears faster and cleaner at early time steps compared to other choices, as shown in~\cref{fig:Teaser}.
We performed an additional experiment called early stopping. 
This is to show that if we stop at early steps, using only \GaussianBlueNoiseText~gives better results than using only Gaussian noise. 
\revision{As shown in Supplemental document Fig. 3,}
the results of using only Gaussian blue noise (second row) stopped at $t=200$ are with sharper details than using only Gaussian noise (first row).
Quantitative evaluation of early stopping can be found in the \SupplementalDoc Sec. 3.
However, as blue noise has no energy in low-frequency region, it is restricting the diffusion process to a limited range of directions. 
For this reason, it becomes difficult to refine the intermediately generated content in later time steps and thus results in worse quality, as shown in~\cref{tab:ablationNoises}.
Instead, our method takes blue noise into account from middle or later time steps, when low-frequency components are already visible and the network is more focusing on refining high-frequency details.

\paragraph{Diffusion with different noise magnitude.}
Since using only \GaussianBlueNoiseText~at all time steps leads to degraded quality, we further conduct an analysis on diffusion at later time steps by explicitly ignoring the early time steps.
We compare Gaussian noise and \GaussianBlueNoiseText~by training a diffusion model (IADB) up to some time steps with certain noise magnitude (e.g., 30\%). 
During the testing phase, ground truth images provided so we are able to compare the denoised images with the ground truth ones.
Running with 100\% of noises, the experiment would fall back to the standard diffusion generative process.
Based on~\cref{fig:celeba_res64_shortcut_diffusion}, by using \GaussianBlueNoiseText we generate more detail- and content-preserving images compared to the one using Gaussian noise.
This indicates that \GaussianBlueNoiseText~is suitable for denoising when low-frequency components become visible.
This is consistent to our idea blending \GaussianBlueNoiseText~from middle or later time steps.

\paragraph{More ablations} 
\revision{
We further compare different $\gamma$ values and a cosine-based scheduler~\cite{nichol2021improved} used for the $\gamma$-scheduler. 
Also, we compare different Gaussian blue noise mask sizes used for padding/tiling. More details can be found in Supplemental document Sec. 3.
}

\section{Conclusion}

We have presented a new method for incorporating correlated noise into deterministic generative diffusion models. Our technique involves using a combination of uncorrelated and correlated noise masks generated using matrix-based methods. By investigating different noise correlation, we have uncovered the intricate relationship between noise characteristics and the quality of generated images. Our findings indicate that high-frequency noise is effective at preserving details but struggles with generating low-frequency components, whereas low-frequency noise hinders the generation of complex details. To achieve optimal image quality, we propose selectively using different types of noise in a time-dependent manner, leveraging the strengths of each noise component. 
To validate the effectiveness of our approach, we conducted extensive experiments using it in conjunction with the well-known method IADB~\cite{heitz2023iterative}. By keeping the training data and optimization hyperparameters consistent, we consistently observed significant improvements in image quality across various datasets. 
\revision{These results demonstrate the superiority of our approach in enhancing image generation capabilities of deterministic diffusion models.
}

\paragraph{Limitations.} 
\revision{
Currently, parameter tuning for our $\gamma$-scheduler depends on the image resolution. It is also computationally intensive to compute Gaussian blue noise masks while extending our approach to higher resolution models.
All data from a specific mini-batch must be on a single GPU for our rectified mapping to function. Further research is needed to expand this to distributed training involving inter-GPU synchronization.
}

\paragraph{Future work.} 
We believe our proposed model will inspire new research directions in designing noise patterns for improving efficiency of generative diffusion models.
An interesting future work would be extending our model to interpolate more than two noises to take into account more different types of noises, such as low-pass and band-pass noises.
This may provide more degree of freedom to improve the training and sampling efficiency of the diffusion models.
Further, we can design more advanced techniques to correlate data samples during training, which is orthogonal to using correlated noise.
Extending our framework (e.g., the time-varying noise model) to \revision{stochastic models~\cite{ho2020denoising, song2020score} and even fewer-step models~\cite{karras2022elucidating, song2023consistency, luo2023latent}} would be another interesting future direction. In this way, our framework can be generalized to state-of-the-art denoising diffusion models.

In terms of applications, we tested our model on 2D unconditional and conditional image generation. 
Interesting future work would include generalizing our model to synthesize other data representation such as video and 3D mesh.

\begin{acks}
We would like to thank the anonymous reviewers for their detailed and constructive comments. 
This project was also supported by Saarbr{\"u}cken Research Center for Visual Computing, Interaction and AI.
\end{acks}

\bibliographystyle{ACM-Reference-Format}
\bibliography{paper}


\newpage
\hbox{}\newpage

\begin{figure}[H]
    \centering

\newcommand{\PlotSingleImage}[1]{%
        \begin{scope}
            \clip (0,0) -- (2.5,0) -- (2.5,2.5) -- (0,2.5) -- cycle;
            \path[fill overzoom image=figures/church_res128_superres_4x/#1] (0,0) rectangle (2.5cm,2.5cm);
        \end{scope}
        \draw (0,0) -- (2.5,0) -- (2.5,2.5) -- (0,2.5) -- cycle;
}

\newcommand{\PlotImageAndCrop}[2]{%
    \begin{scope}
        \clip (0,0) -- (3.2,0) -- (3.2,3.25)-- (0.0,3.25) -- cycle;
        \path[fill overzoom image=figures/noise_visualization/#1] (0,0) rectangle (3.25,3.25);
    \end{scope}
    \begin{scope}
        \clip (2.0,0.1) -- (3.1,0.1) -- (3.1,1.2) -- (2.0,1.2) -- cycle;
        \path[fill overzoom image=figures/noise_visualization/#2] (2.0,0.1) rectangle (3.1cm,1.2cm);
    \end{scope}
}

\newcommand{\TwoColumnFigure}[2]{%
    \begin{tabular}{c@{\;}c@{}}
        \hspace*{-2.5mm}
        \begin{tikzpicture}[scale=0.563]
            \PlotSingleImage{#1}
        \end{tikzpicture}
         & 
         \begin{tikzpicture}[scale=0.563]
            \PlotSingleImage{#2}
        \end{tikzpicture}
    \end{tabular}%
}

\newcommand{\scaleval}{1.12}    
\small
\hspace*{-4mm}
\begin{tabular}{c@{\;}c@{\;}c@{\;}c@{}}
~ &
Input &
IADB &
Ours
\\
&
\begin{tikzpicture}[scale=\scaleval]
    \PlotSingleImage{lowres_gwn_00074.png}
\end{tikzpicture}
&
\begin{tikzpicture}[scale=\scaleval]
    \PlotSingleImage{image_gwn_00074.png}
    \filldraw[white,ultra thick] (1.8, 2.5) circle (0pt) node[anchor=north,rotate=0] 
            {{0.052 (1.0$\times$)}};
\end{tikzpicture}
&
\begin{tikzpicture}[scale=\scaleval]
    \PlotSingleImage{image_gwn2gbn_00074.png}
    \filldraw[white,ultra thick] (1.7, 2.5) circle (0pt) node[anchor=north,rotate=0] 
            {{\textbf{0.043 (0.83$\times$)}}};
\end{tikzpicture}
\\[-0.4mm]
&
\begin{tikzpicture}[scale=\scaleval]
    \PlotSingleImage{lowres_gwn_00104.png}
\end{tikzpicture}
&
\begin{tikzpicture}[scale=\scaleval]
    \PlotSingleImage{image_gwn_00104.png}
    \filldraw[black,ultra thick] (1.8, 2.5) circle (0pt) node[anchor=north,rotate=0] 
            {{0.040 (1.0$\times$)}};
\end{tikzpicture}
&
\begin{tikzpicture}[scale=\scaleval]
    \PlotSingleImage{image_gwn2gbn_00104.png}
    \filldraw[black,ultra thick] (1.7, 2.5) circle (0pt) node[anchor=north,rotate=0] 
            {{\textbf{0.034 (0.85$\times$)}}};
\end{tikzpicture}
\\[-0.4mm]
&
\begin{tikzpicture}[scale=\scaleval]
    \PlotSingleImage{lowres_gwn_00278.png}
\end{tikzpicture}
&
\begin{tikzpicture}[scale=\scaleval]
    \PlotSingleImage{image_gwn_00278.png}
    \filldraw[white,ultra thick] (1.8, 2.5) circle (0pt) node[anchor=north,rotate=0] 
            {{0.105 (1.0$\times$)}};
\end{tikzpicture}
&
\begin{tikzpicture}[scale=\scaleval]
    \PlotSingleImage{image_gwn2gbn_00278.png}
    \filldraw[white,ultra thick] (1.7, 2.5) circle (0pt) node[anchor=north,rotate=0] 
            {{\textbf{0.098 (0.93$\times$)}}};
\end{tikzpicture}
\\[-0.4mm]
&
\begin{tikzpicture}[scale=\scaleval]
    \PlotSingleImage{lowres_gwn_00389.png}
\end{tikzpicture}
&
\begin{tikzpicture}[scale=\scaleval]
    \PlotSingleImage{image_gwn_00389.png}
    \filldraw[white,ultra thick] (1.8, 2.5) circle (0pt) node[anchor=north,rotate=0] 
            {{0.047 (1.0$\times$)}};
\end{tikzpicture}
&
\begin{tikzpicture}[scale=\scaleval]
    \PlotSingleImage{image_gwn2gbn_00389.png}
    \filldraw[white,ultra thick] (1.7, 2.5) circle (0pt) node[anchor=north,rotate=0] 
            {{\textbf{0.038 (0.82$\times$)}}};
\end{tikzpicture}
\end{tabular}
    \vspace{-2.5mm} 
    \caption{
        Image super-resolution comparisons between IADB (SSIM/PSNR=0.57/19.46) and Ours (SSIM/PSNR=\textbf{0.59}/\textbf{20.00}) on LSUN-Church ($32^2 \rightarrow 128^2$).
        The mean squared error w.r.t the reference is visible in the upper corner with the relative error to IADB. Our method achieves lower error and more plausible details with less hallucination.
    }
    \label{fig:celeba_res128_superres}
\end{figure}

\begin{figure}[h]
    \centering

\newcommand{\PlotSingleImage}[1]{%
        \begin{scope}
            \clip (0,0) -- (2.5,0) -- (2.5,2.5) -- (0,2.5) -- cycle;
            \path[fill overzoom image=figures/cat_res128_iadb_gwn_gbn_grn/#1] (0,0) rectangle (2.5cm,2.5cm);
        \end{scope}
        \draw (0,0) -- (2.5,0) -- (2.5,2.5) -- (0,2.5) -- cycle;
}

\newcommand{\TwoColumnFigure}[2]{%
    \begin{tabular}{c@{\;}c@{}}
        \hspace*{-2.5mm}
        \begin{tikzpicture}[scale=0.563]
            \PlotSingleImage{#1}
        \end{tikzpicture}
         & 
         \begin{tikzpicture}[scale=0.563]
            \PlotSingleImage{#2}
        \end{tikzpicture}
    \end{tabular}%
}

\newcommand{\scaleval}{0.81}    
\small
\hspace*{-4mm}
\begin{tabular}{c@{\;}c@{\;}c@{\;}c@{\;}c@{}}
~ &
\\
\rotatebox{90}{\hspace{0.15cm} \scriptsize Ours (white+red)}
&
\begin{tikzpicture}[scale=\scaleval]
    \PlotSingleImage{gwn2grn_img03600_step250.png}
\end{tikzpicture}
&
\begin{tikzpicture}[scale=\scaleval]
    \PlotSingleImage{gwn2grn_img03800_step250.png}
\end{tikzpicture}
&
\begin{tikzpicture}[scale=\scaleval]
    \PlotSingleImage{gwn2grn_img04600_step250.png}
\end{tikzpicture}
&
\begin{tikzpicture}[scale=\scaleval]
    \PlotSingleImage{gwn2grn_img07800_step250.png}
\end{tikzpicture}
\\[-0.4mm]
\rotatebox{90}{\hspace{0.15cm} \scriptsize Ours (white+blue)}
&
\begin{tikzpicture}[scale=\scaleval]
    \PlotSingleImage{gwn2gbn_img03600_step250.png}
\end{tikzpicture}
&
\begin{tikzpicture}[scale=\scaleval]
    \PlotSingleImage{gwn2gbn_img03800_step250.png}
\end{tikzpicture}
&
\begin{tikzpicture}[scale=\scaleval]
    \PlotSingleImage{gwn2gbn_img04600_step250.png}
\end{tikzpicture}
&
\begin{tikzpicture}[scale=\scaleval]
    \PlotSingleImage{gwn2gbn_img07800_step250.png}
\end{tikzpicture}
\\[-0.4mm]
\end{tabular} 
    \vspace{-2.5mm} 
    \caption{
        Qualitative image generation comparisons between Gaussian red noise and Gaussian blue noise using our method on AFHQ-Cat ($128^2$). 
        Our method with Gaussian blue noise creates more high-frequency details while using Gaussian red noise introduces visible artifacts. 
    }
    \label{fig:cat_res128_compare_gwn_gbn_grn}
\end{figure}

\begin{figure}[H]
    \centering

\newcommand{\PlotSingleImage}[1]{%
        \begin{scope}
            \clip (0,0) -- (2.5,0) -- (2.5,2.5) -- (0,2.5) -- cycle;
            \path[fill overzoom image=figures/celeba_res64_shortcut_diffusion/#1] (0,0) rectangle (2.5cm,2.5cm);
        \end{scope}
        \draw (0,0) -- (2.5,0) -- (2.5,2.5) -- (0,2.5) -- cycle;
}

\newcommand{\TwoColumnFigure}[2]{%
    \begin{tabular}{c@{\;}c@{}}
        \hspace*{-2.5mm}
        \begin{tikzpicture}[scale=0.563]
            \PlotSingleImage{#1}
        \end{tikzpicture}
         & 
         \begin{tikzpicture}[scale=0.563]
            \PlotSingleImage{#2}
        \end{tikzpicture}
    \end{tabular}%
}

\newcommand{\scaleval}{0.65}
\small
\hspace*{-4mm}
\begin{tabular}{c@{\;}c@{\;}c@{\;}c@{\;}c@{\;}c@{\;}c@{\;}c@{\;}c@{}}
~ &
$100\%$ & 
$70\%$ &
$50\%$ &
$30\%$ &
Reference
\\
\rotatebox{90}{\hspace{0.15cm} \scriptsize Random noise}
&
\begin{tikzpicture}[scale=\scaleval]
\PlotSingleImage{gaussian_alpha0.00_img00000.png}
\end{tikzpicture}
&
\begin{tikzpicture}[scale=\scaleval]
\PlotSingleImage{gaussian_alpha0.30_img00000.png}
\end{tikzpicture}
&
\begin{tikzpicture}[scale=\scaleval]
\PlotSingleImage{gaussian_alpha0.50_img00000.png}
\end{tikzpicture}
&
\begin{tikzpicture}[scale=\scaleval]
\PlotSingleImage{gaussian_alpha0.70_img00000.png}
\end{tikzpicture}
&
\begin{tikzpicture}[scale=\scaleval]
\PlotSingleImage{0025001.png}
\end{tikzpicture}
\\[-0.4mm]
\rotatebox{90}{\hspace{0.25cm} \scriptsize Blue noise}
&
\begin{tikzpicture}[scale=\scaleval]
\PlotSingleImage{gaussianBN_alpha0.00_img00000.png}
\end{tikzpicture}
&
\begin{tikzpicture}[scale=\scaleval]
\PlotSingleImage{gaussianBN_alpha0.30_img00000.png}
\end{tikzpicture}
&
\begin{tikzpicture}[scale=\scaleval]
\PlotSingleImage{gaussianBN_alpha0.50_img00000.png}
\end{tikzpicture}
&
\begin{tikzpicture}[scale=\scaleval]
\PlotSingleImage{gaussianBN_alpha0.70_img00000.png}
\end{tikzpicture}
&
\begin{tikzpicture}[scale=\scaleval]
\PlotSingleImage{0025001.png}
\end{tikzpicture}
\\[0.5mm]
\rotatebox{90}{\hspace{0.15cm} \scriptsize Random noise}
&
\begin{tikzpicture}[scale=\scaleval]
\PlotSingleImage{gaussian_img00972_alpha0.00.png}
\end{tikzpicture}
&
\begin{tikzpicture}[scale=\scaleval]
\PlotSingleImage{gaussian_img00972_alpha0.30.png}
\end{tikzpicture}
&
\begin{tikzpicture}[scale=\scaleval]
\PlotSingleImage{gaussian_img00972_alpha0.50.png}
\end{tikzpicture}
&
\begin{tikzpicture}[scale=\scaleval]
\PlotSingleImage{gaussian_img00972_alpha0.70.png}
\end{tikzpicture}
&
\begin{tikzpicture}[scale=\scaleval]
\PlotSingleImage{0025973.png}
\end{tikzpicture}
\\[-0.4mm]
\rotatebox{90}{\hspace{0.25cm} \scriptsize Blue noise}
&
\begin{tikzpicture}[scale=\scaleval]
\PlotSingleImage{gaussianBN_img00972_alpha0.00.png}
\end{tikzpicture}
&
\begin{tikzpicture}[scale=\scaleval]
\PlotSingleImage{gaussianBN_img00972_alpha0.30.png}
\end{tikzpicture}
&
\begin{tikzpicture}[scale=\scaleval]
\PlotSingleImage{gaussianBN_img00972_alpha0.50.png}
\end{tikzpicture}
&
\begin{tikzpicture}[scale=\scaleval]
\PlotSingleImage{gaussianBN_img00972_alpha0.70.png}
\end{tikzpicture}
&
\begin{tikzpicture}[scale=\scaleval]
\PlotSingleImage{0025973.png}
\end{tikzpicture}
\\[-0.4mm]
\end{tabular} 
    \vspace{-2.5mm}   
    \caption{
        Evaluating the impact of noise magnitude on detail enhancement. 
        Our Gaussian blue noise method better preserves fine details even with increased noise magnitude, while maintaining the integrity of the content.
        With $100\%$ noise, both models fall back to full generative process.
    }
    \label{fig:celeba_res64_shortcut_diffusion}
\end{figure}

\begin{figure}[h]
    \centering

\newcommand{\PlotSingleImage}[1]{%
        \begin{scope}
            \clip (0,0) -- (2.5,0) -- (2.5,2.5) -- (0,2.5) -- cycle;
            \path[fill overzoom image=figures/cat_res512_iadb/#1] (0,0) rectangle (2.5cm,2.5cm);
        \end{scope}
        \draw (0,0) -- (2.5,0) -- (2.5,2.5) -- (0,2.5) -- cycle;
}

\newcommand{\TwoColumnFigure}[2]{%
    \begin{tabular}{c@{\;}c@{}}
        \hspace*{-2.5mm}
        \begin{tikzpicture}[scale=0.563]
            \PlotSingleImage{#1}
        \end{tikzpicture}
         & 
         \begin{tikzpicture}[scale=0.563]
            \PlotSingleImage{#2}
        \end{tikzpicture}
    \end{tabular}%
}

\newcommand{\scaleval}{0.64}    
\small
\hspace*{-2.5mm}  
\begin{tabular}{c@{\;}c@{\;}c@{\;}c@{\;}c@{\;}c@{}}
~ &
\\
\rotatebox{90}{\hspace{0.2cm} \scriptsize LDM (IADB)}
&
\begin{tikzpicture}[scale=\scaleval]
    \PlotSingleImage{iadb_gwn_00003.png}
\end{tikzpicture}
&
\begin{tikzpicture}[scale=\scaleval]
    \PlotSingleImage{iadb_gwn_00008.png}
\end{tikzpicture}
&
\begin{tikzpicture}[scale=\scaleval]
    \PlotSingleImage{iadb_gwn_00049.png}
\end{tikzpicture}
&
\begin{tikzpicture}[scale=\scaleval]
    \PlotSingleImage{iadb_gwn_00057.png}
\end{tikzpicture}
&
\begin{tikzpicture}[scale=\scaleval]
    \PlotSingleImage{iadb_gwn_00032.png}
\end{tikzpicture}
\\[-0.4mm]
\rotatebox{90}{\hspace{0.2cm} \scriptsize LDM (Ours)}
&
\begin{tikzpicture}[scale=\scaleval]
    \PlotSingleImage{iadb_gwn2gbn_00003.png}
\end{tikzpicture}
&
\begin{tikzpicture}[scale=\scaleval]
    \PlotSingleImage{iadb_gwn2gbn_00008.png}
\end{tikzpicture}
&
\begin{tikzpicture}[scale=\scaleval]
    \PlotSingleImage{iadb_gwn2gbn_00049.png}
\end{tikzpicture}
&
\begin{tikzpicture}[scale=\scaleval]
    \PlotSingleImage{iadb_gwn2gbn_00057.png}
\end{tikzpicture}
&
\begin{tikzpicture}[scale=\scaleval]
    \PlotSingleImage{iadb_gwn2gbn_00032.png}
\end{tikzpicture}
\\[-0.4mm]
\end{tabular} 
    \vspace{-2.5mm} 
    \caption{
        \revision{Latent diffusion model (LDM) based high-res image generation using IADB and Ours on AFHQ-Cat ($512^2$). IADB introduces more artifacts around the eye regions and has worse FID ($12.19>\textbf{11.45}$) compared with Ours.
        }
    }
    \label{fig:cat_res512_compare_gwn_gbn}
\end{figure}

\begin{figure}[h]
    \centering

\newcommand{\PlotSingleImage}[1]{%
        \begin{scope}
            \clip (0,0) -- (2.5,0) -- (2.5,2.5) -- (0,2.5) -- cycle;
            \path[fill overzoom image=figures/noise_visualization/#1] (0,0) rectangle (2.5cm,2.5cm);
        \end{scope}
        \draw (0,0) -- (2.5,0) -- (2.5,2.5) -- (0,2.5) -- cycle;
}

\newcommand{\PlotImageAndCrop}[2]{%
    \begin{scope}
        \clip (0,0) -- (3.2,0) -- (3.2,3.25)-- (0.0,3.25) -- cycle;
        \path[fill overzoom image=figures/noise_visualization/#1] (0,0) rectangle (3.25,3.25);
    \end{scope}
    \begin{scope}
        \clip (2.0,0.1) -- (3.1,0.1) -- (3.1,1.2) -- (2.0,1.2) -- cycle;
        \path[fill overzoom image=figures/noise_visualization/#2] (2.0,0.1) rectangle (3.1cm,1.2cm);
    \end{scope}
}

\newcommand{\TwoColumnFigure}[2]{%
    \begin{tabular}{c@{\;}c@{}}
        \hspace*{-2.5mm}
        \begin{tikzpicture}[scale=0.563]
            \PlotSingleImage{#1}
        \end{tikzpicture}
         & 
         \begin{tikzpicture}[scale=0.563]
            \PlotSingleImage{#2}
        \end{tikzpicture}
    \end{tabular}%
}

\newcommand{\scaleval}{0.89}    
\small
\hspace*{-4mm}
\begin{tabular}{c@{\;}c@{\;}c@{\;}c@{}}
~ &
Gaussian noise &
Gaussian blue noise &
Gaussian red noise
\\
&
\begin{tikzpicture}[scale=\scaleval]
    \PlotImageAndCrop{wn_res64.png}{wn_res64_ps.png}
\end{tikzpicture}
&
\begin{tikzpicture}[scale=\scaleval]
    \PlotImageAndCrop{bn_res64.png}{bn_res64_ps.png}
\end{tikzpicture}
&
\begin{tikzpicture}[scale=\scaleval]
    \PlotImageAndCrop{rn_res64.png}{rn_res64_ps.png}
\end{tikzpicture}
\\[-0.4mm]
\end{tabular} 
    \vspace{-2.5mm} 
    \caption{
        Visualization of Gaussian noise, Gaussian blue noise and Gaussian red noise at resolution $64^2$. According Power spectrum is shown in the bottom right corner. 
         \revision{While Gaussian blue noise show mostly high frequency variations, Gaussian red noise shows only low frequencies.}
    }
    \label{fig:gwn_gbw_grn}
\end{figure}

\begin{figure*}[t!]
    \centering

\newcommand{\PlotSingleImage}[1]{%
        \begin{scope}
            \clip (0,0) -- (2.5,0) -- (2.5,2.5) -- (0,2.5) -- cycle;
            \path[fill overzoom image=figures/#1] (0,0) rectangle (2.5cm,2.5cm);
        \end{scope}
        \draw (0,0) -- (2.5,0) -- (2.5,2.5) -- (0,2.5) -- cycle;
        
}

\newcommand{\PlotSingleImageWithLine}[1]{%
        \begin{scope}
            \clip (0,0) -- (2.5,0) -- (2.5,2.5) -- (0,2.5) -- cycle;
            \path[fill overzoom image=figures/#1] (0,0) rectangle (2.5cm,2.5cm);
        \end{scope}
        \draw (0,0) -- (2.5,0) -- (2.5,2.5) -- (0,2.5) -- cycle;
        \draw[dashed] (0, -0.13) -- (2.5, -0.13);
}

\newcommand{\TwoColumnFigure}[2]{%
    \begin{tabular}{c@{\;}c@{}}
        \hspace*{-2.5mm}
        \begin{tikzpicture}[scale=0.563]
            \PlotSingleImage{#1}
        \end{tikzpicture}
         & 
         \begin{tikzpicture}[scale=0.563]
            \PlotSingleImage{#2}
        \end{tikzpicture}
    \end{tabular}%
}
\newcommand\scalevalue{0.65}
\small
\hspace*{-4mm}
\begin{tabular}{c@{\;}c@{}}
\begin{tabular}{c@{\;}c@{\;}c@{\;}c@{\;}c@{\;}c@{}}
\rotatebox{90}{\hspace{0.5cm} \scriptsize DDIM}
&
\begin{tikzpicture}[scale=\scalevalue]
\PlotSingleImage{church_res64_iadb/ddim_img02000_step0.png}
\end{tikzpicture}
&
\begin{tikzpicture}[scale=\scalevalue]
\PlotSingleImage{church_res64_iadb/ddim_img02000_step100.png}
\end{tikzpicture}
&
\begin{tikzpicture}[scale=\scalevalue]
\PlotSingleImage{church_res64_iadb/ddim_img02000_step175.png}
\end{tikzpicture}
&
\begin{tikzpicture}[scale=\scalevalue]
\PlotSingleImage{church_res64_iadb/ddim_img02000_step225.png}
\end{tikzpicture}
&
\begin{tikzpicture}[scale=\scalevalue]
\PlotSingleImage{church_res64_iadb/ddim_img02000_step250.png}
\end{tikzpicture}
\\[-0.4mm]
\rotatebox{90}{\hspace{0.5cm} \scriptsize IADB}
&
\begin{tikzpicture}[scale=\scalevalue]
\PlotSingleImage{church_res64_iadb/gwn_img02000_step0.png}
\end{tikzpicture}
&
\begin{tikzpicture}[scale=\scalevalue]
\PlotSingleImage{church_res64_iadb/gwn_img02000_step100.png}
\end{tikzpicture}
&
\begin{tikzpicture}[scale=\scalevalue]
\PlotSingleImage{church_res64_iadb/gwn_img02000_step175.png}
\end{tikzpicture}
&
\begin{tikzpicture}[scale=\scalevalue]
\PlotSingleImage{church_res64_iadb/gwn_img02000_step225.png}
\end{tikzpicture}
&
\begin{tikzpicture}[scale=\scalevalue]
\PlotSingleImage{church_res64_iadb/gwn_img02000_step250.png}
\end{tikzpicture}
\\[-0.4mm]
\rotatebox{90}{\hspace{0.5cm} \scriptsize Ours}
&
\begin{tikzpicture}[scale=\scalevalue]
\PlotSingleImageWithLine{church_res64_iadb/gwn2gbn_img02000_step0.png}
\end{tikzpicture}
&
\begin{tikzpicture}[scale=\scalevalue]
\PlotSingleImageWithLine{church_res64_iadb//gwn2gbn_img02000_step100.png}
\end{tikzpicture}
&
\begin{tikzpicture}[scale=\scalevalue]
\PlotSingleImageWithLine{church_res64_iadb/gwn2gbn_img02000_step175.png}
\end{tikzpicture}
&
\begin{tikzpicture}[scale=\scalevalue]
\PlotSingleImageWithLine{church_res64_iadb/gwn2gbn_img02000_step225.png}
\end{tikzpicture}
&
\begin{tikzpicture}[scale=\scalevalue]
\PlotSingleImageWithLine{church_res64_iadb/gwn2gbn_img02000_step250.png}
\end{tikzpicture}
\\[-0.4mm]
\rotatebox{90}{\hspace{0.5cm} \scriptsize DDIM}
&
\begin{tikzpicture}[scale=\scalevalue]
\PlotSingleImage{church_res64_iadb/ddim_img11500_step0.png}
\end{tikzpicture}
&
\begin{tikzpicture}[scale=\scalevalue]
\PlotSingleImage{church_res64_iadb/ddim_img11500_step100.png}
\end{tikzpicture}
&
\begin{tikzpicture}[scale=\scalevalue]
\PlotSingleImage{church_res64_iadb/ddim_img11500_step175.png}
\end{tikzpicture}
&
\begin{tikzpicture}[scale=\scalevalue]
\PlotSingleImage{church_res64_iadb/ddim_img11500_step225.png}
\end{tikzpicture}
&
\begin{tikzpicture}[scale=\scalevalue]
\PlotSingleImage{church_res64_iadb/ddim_img11500_step250.png}
\end{tikzpicture}
\\[-0.4mm]
\rotatebox{90}{\hspace{0.5cm} \scriptsize IADB}
&
\begin{tikzpicture}[scale=\scalevalue]
\PlotSingleImage{church_res64_iadb/gwn_img11500_step0.png}
\end{tikzpicture}
&
\begin{tikzpicture}[scale=\scalevalue]
\PlotSingleImage{church_res64_iadb/gwn_img11500_step100.png}
\end{tikzpicture}
&
\begin{tikzpicture}[scale=\scalevalue]
\PlotSingleImage{church_res64_iadb/gwn_img11500_step175.png}
\end{tikzpicture}
&
\begin{tikzpicture}[scale=\scalevalue]
\PlotSingleImage{church_res64_iadb/gwn_img11500_step225.png}
\end{tikzpicture}
&
\begin{tikzpicture}[scale=\scalevalue]
\PlotSingleImage{church_res64_iadb/gwn_img11500_step250.png}
\end{tikzpicture}
\\[-0.4mm]
\rotatebox{90}{\hspace{0.5cm} \scriptsize Ours}
&
\begin{tikzpicture}[scale=\scalevalue]
\PlotSingleImage{church_res64_iadb/gwn2gbn_img11500_step0.png}
\end{tikzpicture}
&
\begin{tikzpicture}[scale=\scalevalue]
\PlotSingleImage{church_res64_iadb//gwn2gbn_img11500_step100.png}
\end{tikzpicture}
&
\begin{tikzpicture}[scale=\scalevalue]
\PlotSingleImage{church_res64_iadb/gwn2gbn_img11500_step175.png}
\end{tikzpicture}
&
\begin{tikzpicture}[scale=\scalevalue]
\PlotSingleImage{church_res64_iadb/gwn2gbn_img11500_step225.png}
\end{tikzpicture}
&
\begin{tikzpicture}[scale=\scalevalue]
\PlotSingleImage{church_res64_iadb/gwn2gbn_img11500_step250.png}
\end{tikzpicture}
\\[-0.4mm]
~ &
$t=250$ & 
$t=100$ &
$t=75$ &
$t=25$ &
$t=0$
\\[-0.4mm]
\end{tabular}
&
\begin{tabular}{c@{\;}c@{\;}c@{\;}c@{\;}c@{\;}c@{}}
\rotatebox{90}{\hspace{0.5cm} \scriptsize DDIM}
&
\begin{tikzpicture}[scale=\scalevalue]
\PlotSingleImage{cat_res64_iadb/ddim_img02000_step0.png}
\end{tikzpicture}
&
\begin{tikzpicture}[scale=\scalevalue]
\PlotSingleImage{cat_res64_iadb/ddim_img02000_step100.png}
\end{tikzpicture}
&
\begin{tikzpicture}[scale=\scalevalue]
\PlotSingleImage{cat_res64_iadb/ddim_img02000_step175.png}
\end{tikzpicture}
&
\begin{tikzpicture}[scale=\scalevalue]
\PlotSingleImage{cat_res64_iadb/ddim_img02000_step225.png}
\end{tikzpicture}
&
\begin{tikzpicture}[scale=\scalevalue]
\PlotSingleImage{cat_res64_iadb/ddim_img02000_step250.png}
\end{tikzpicture}
\\[-0.4mm]
\rotatebox{90}{\hspace{0.5cm} \scriptsize IADB}
&
\begin{tikzpicture}[scale=\scalevalue]
\PlotSingleImage{cat_res64_iadb/gwn_img02000_step0.png}
\end{tikzpicture}
&
\begin{tikzpicture}[scale=\scalevalue]
\PlotSingleImage{cat_res64_iadb/gwn_img02000_step100.png}
\end{tikzpicture}
&
\begin{tikzpicture}[scale=\scalevalue]
\PlotSingleImage{cat_res64_iadb/gwn_img02000_step175.png}
\end{tikzpicture}
&
\begin{tikzpicture}[scale=\scalevalue]
\PlotSingleImage{cat_res64_iadb/gwn_img02000_step225.png}
\end{tikzpicture}
&
\begin{tikzpicture}[scale=\scalevalue]
\PlotSingleImage{cat_res64_iadb/gwn_img02000_step250.png}
\end{tikzpicture}
\\[-0.4mm]
\rotatebox{90}{\hspace{0.5cm} \scriptsize Ours}
&
\begin{tikzpicture}[scale=\scalevalue]
\PlotSingleImageWithLine{cat_res64_iadb/gwn2gbn_img02000_step0.png}
\end{tikzpicture}
&
\begin{tikzpicture}[scale=\scalevalue]
\PlotSingleImageWithLine{cat_res64_iadb/gwn2gbn_img02000_step100.png}
\end{tikzpicture}
&
\begin{tikzpicture}[scale=\scalevalue]
\PlotSingleImageWithLine{cat_res64_iadb/gwn2gbn_img02000_step175.png}
\end{tikzpicture}
&
\begin{tikzpicture}[scale=\scalevalue]
\PlotSingleImageWithLine{cat_res64_iadb/gwn2gbn_img02000_step225.png}
\end{tikzpicture}
&
\begin{tikzpicture}[scale=\scalevalue]
\PlotSingleImageWithLine{cat_res64_iadb/gwn2gbn_img02000_step250.png}
\end{tikzpicture}
\\[-0.4mm]
\rotatebox{90}{\hspace{0.5cm} \scriptsize DDIM}
&
\begin{tikzpicture}[scale=\scalevalue]
\PlotSingleImage{celeba_res64_iadb/ddim_img18500_step0.png}
\end{tikzpicture}
&
\begin{tikzpicture}[scale=\scalevalue]
\PlotSingleImage{celeba_res64_iadb/ddim_img18500_step100.png}
\end{tikzpicture}
&
\begin{tikzpicture}[scale=\scalevalue]
\PlotSingleImage{celeba_res64_iadb/ddim_img18500_step175.png}
\end{tikzpicture}
&
\begin{tikzpicture}[scale=\scalevalue]
\PlotSingleImage{celeba_res64_iadb/ddim_img18500_step225.png}
\end{tikzpicture}
&
\begin{tikzpicture}[scale=\scalevalue]
\PlotSingleImage{celeba_res64_iadb/ddim_img18500_step250.png}
\end{tikzpicture}
\\[-0.4mm]
\rotatebox{90}{\hspace{0.5cm} \scriptsize IADB}
&
\begin{tikzpicture}[scale=\scalevalue]
\PlotSingleImage{celeba_res64_iadb/gwn_img18500_step0.png}
\end{tikzpicture}
&
\begin{tikzpicture}[scale=\scalevalue]
\PlotSingleImage{celeba_res64_iadb/gwn_img18500_step100.png}
\end{tikzpicture}
&
\begin{tikzpicture}[scale=\scalevalue]
\PlotSingleImage{celeba_res64_iadb/gwn_img18500_step175.png}
\end{tikzpicture}
&
\begin{tikzpicture}[scale=\scalevalue]
\PlotSingleImage{celeba_res64_iadb/gwn_img18500_step225.png}
\end{tikzpicture}
&
\begin{tikzpicture}[scale=\scalevalue]
\PlotSingleImage{celeba_res64_iadb/gwn_img18500_step250.png}
\end{tikzpicture}
\\[-0.4mm]
\rotatebox{90}{\hspace{0.5cm} \scriptsize Ours}
&
\begin{tikzpicture}[scale=\scalevalue]
\PlotSingleImage{celeba_res64_iadb/gwn2gbn_img18500_step0.png}
\end{tikzpicture}
&
\begin{tikzpicture}[scale=\scalevalue]
\PlotSingleImage{celeba_res64_iadb/gwn2gbn_img18500_step100.png}
\end{tikzpicture}
&
\begin{tikzpicture}[scale=\scalevalue]
\PlotSingleImage{celeba_res64_iadb/gwn2gbn_img18500_step175.png}
\end{tikzpicture}
&
\begin{tikzpicture}[scale=\scalevalue]
\PlotSingleImage{celeba_res64_iadb/gwn2gbn_img18500_step225.png}
\end{tikzpicture}
&
\begin{tikzpicture}[scale=\scalevalue]
\PlotSingleImage{celeba_res64_iadb/gwn2gbn_img18500_step250.png}
\end{tikzpicture}
\\[-0.4mm]
~ &
$t=250$ & 
$t=100$ &
$t=75$ &
$t=25$ &
$t=0$
\\[-0.4mm]
\end{tabular}
\end{tabular} 
    \vspace{-2.5mm} 
    \caption{
        Comparisons of image generation using DDIM, IADB and Ours trained on LSUN-Church ($64^2$), AFHQ-Cat ($64^2$) and CelebA ($64^2$) datasets. 
        For each example all methods start the diffusion from the same noise. 
        In all cases our method achieve the highest quality result with more realistic images. Quality in details generation can be seen in the windows and doors of the buildings. By looking at the noise at different time steps the evolution from random to blue noise is visible for our method.
    }
    \label{fig:church_res64_sequence_gwn_gbn}
\end{figure*}

\begin{figure*}[t!]
    \centering

\newcommand{\PlotSingleImage}[1]{%
        \begin{scope}
            \clip (0,0) -- (2.5,0) -- (2.5,2.5) -- (0,2.5) -- cycle;
            \path[fill overzoom image=figures/#1] (0,0) rectangle (2.5cm,2.5cm);
        \end{scope}
        \draw (0,0) -- (2.5,0) -- (2.5,2.5) -- (0,2.5) -- cycle;
        
}

\newcommand{\PlotSingleImageWithLine}[1]{%
        \begin{scope}
            \clip (0,0) -- (2.5,0) -- (2.5,2.5) -- (0,2.5) -- cycle;
            \path[fill overzoom image=figures/#1] (0,0) rectangle (2.5cm,2.5cm);
        \end{scope}
        \draw (0,0) -- (2.5,0) -- (2.5,2.5) -- (0,2.5) -- cycle;
        \draw[dashed] (0, -0.13) -- (2.5, -0.13);
}

\newcommand{\TwoColumnFigure}[2]{%
    \begin{tabular}{c@{\;}c@{}}
        \hspace*{-2.5mm}
        \begin{tikzpicture}[scale=0.563]
            \PlotSingleImage{#1}
        \end{tikzpicture}
         & 
         \begin{tikzpicture}[scale=0.563]
            \PlotSingleImage{#2}
        \end{tikzpicture}
    \end{tabular}%
}
\newcommand\scalevalue{0.65}
\small
\hspace*{-4mm}
\begin{tabular}{c@{\;}c@{}}
\begin{tabular}{c@{\;}c@{\;}c@{\;}c@{\;}c@{\;}c@{}}
\rotatebox{90}{\hspace{0.5cm} \scriptsize DDIM}
&
\begin{tikzpicture}[scale=\scalevalue]
\PlotSingleImage{celeba_res128_iadb/ddim_img02000_step0.png}
\end{tikzpicture}
&
\begin{tikzpicture}[scale=\scalevalue]
\PlotSingleImage{celeba_res128_iadb/ddim_img02000_step100.png}
\end{tikzpicture}
&
\begin{tikzpicture}[scale=\scalevalue]
\PlotSingleImage{celeba_res128_iadb/ddim_img02000_step175.png}
\end{tikzpicture}
&
\begin{tikzpicture}[scale=\scalevalue]
\PlotSingleImage{celeba_res128_iadb/ddim_img02000_step225.png}
\end{tikzpicture}
&
\begin{tikzpicture}[scale=\scalevalue]
\PlotSingleImage{celeba_res128_iadb/ddim_img02000_step250.png}
\end{tikzpicture}
\\[-0.4mm]
\rotatebox{90}{\hspace{0.5cm} \scriptsize IADB}
&
\begin{tikzpicture}[scale=\scalevalue]
\PlotSingleImage{celeba_res128_iadb/gwn_img02000_step0.png}
\end{tikzpicture}
&
\begin{tikzpicture}[scale=\scalevalue]
\PlotSingleImage{celeba_res128_iadb/gwn_img02000_step100.png}
\end{tikzpicture}
&
\begin{tikzpicture}[scale=\scalevalue]
\PlotSingleImage{celeba_res128_iadb/gwn_img02000_step175.png}
\end{tikzpicture}
&
\begin{tikzpicture}[scale=\scalevalue]
\PlotSingleImage{celeba_res128_iadb/gwn_img02000_step225.png}
\end{tikzpicture}
&
\begin{tikzpicture}[scale=\scalevalue]
\PlotSingleImage{celeba_res128_iadb/gwn_img02000_step250.png}
\end{tikzpicture}
\\[-0.4mm]
\rotatebox{90}{\hspace{0.5cm} \scriptsize Ours}
&
\begin{tikzpicture}[scale=\scalevalue]
\PlotSingleImage{celeba_res128_iadb/gwn2gbn_img02000_step0.png}
\end{tikzpicture}
&
\begin{tikzpicture}[scale=\scalevalue]
\PlotSingleImage{celeba_res128_iadb/gwn2gbn_img02000_step100.png}
\end{tikzpicture}
&
\begin{tikzpicture}[scale=\scalevalue]
\PlotSingleImage{celeba_res128_iadb/gwn2gbn_img02000_step175.png}
\end{tikzpicture}
&
\begin{tikzpicture}[scale=\scalevalue]
\PlotSingleImage{celeba_res128_iadb/gwn2gbn_img02000_step225.png}
\end{tikzpicture}
&
\begin{tikzpicture}[scale=\scalevalue]
\PlotSingleImage{celeba_res128_iadb/gwn2gbn_img02000_step250.png}
\end{tikzpicture}
\\[-0.4mm]
~ &
$t=250$ & 
$t=100$ &
$t=75$ &
$t=25$ &
$t=0$
\\[-0.4mm]
\end{tabular}
&
\begin{tabular}{c@{\;}c@{\;}c@{\;}c@{\;}c@{\;}c@{}}
\rotatebox{90}{\hspace{0.5cm} \scriptsize DDIM}
&
\begin{tikzpicture}[scale=\scalevalue]
\PlotSingleImage{cat_res128_iadb/ddim_img10400_step0.png}
\end{tikzpicture}
&
\begin{tikzpicture}[scale=\scalevalue]
\PlotSingleImage{cat_res128_iadb/ddim_img10400_step100.png}
\end{tikzpicture}
&
\begin{tikzpicture}[scale=\scalevalue]
\PlotSingleImage{cat_res128_iadb/ddim_img10400_step175.png}
\end{tikzpicture}
&
\begin{tikzpicture}[scale=\scalevalue]
\PlotSingleImage{cat_res128_iadb/ddim_img10400_step225.png}
\end{tikzpicture}
&
\begin{tikzpicture}[scale=\scalevalue]
\PlotSingleImage{cat_res128_iadb/ddim_img10400_step250.png}
\end{tikzpicture}
\\[-0.4mm]
\rotatebox{90}{\hspace{0.5cm} \scriptsize IADB}
&
\begin{tikzpicture}[scale=\scalevalue]
\PlotSingleImage{cat_res128_iadb/gwn_img10400_step0.png}
\end{tikzpicture}
&
\begin{tikzpicture}[scale=\scalevalue]
\PlotSingleImage{cat_res128_iadb/gwn_img10400_step100.png}
\end{tikzpicture}
&
\begin{tikzpicture}[scale=\scalevalue]
\PlotSingleImage{cat_res128_iadb/gwn_img10400_step175.png}
\end{tikzpicture}
&
\begin{tikzpicture}[scale=\scalevalue]
\PlotSingleImage{cat_res128_iadb/gwn_img10400_step225.png}
\end{tikzpicture}
&
\begin{tikzpicture}[scale=\scalevalue]
\PlotSingleImage{cat_res128_iadb/gwn_img10400_step250.png}
\end{tikzpicture}
\\[-0.4mm]
\rotatebox{90}{\hspace{0.5cm} \scriptsize Ours}
&
\begin{tikzpicture}[scale=\scalevalue]
\PlotSingleImage{cat_res128_iadb/gwn2gbn_img10400_step0.png}
\end{tikzpicture}
&
\begin{tikzpicture}[scale=\scalevalue]
\PlotSingleImage{cat_res128_iadb/gwn2gbn_img10400_step100.png}
\end{tikzpicture}
&
\begin{tikzpicture}[scale=\scalevalue]
\PlotSingleImage{cat_res128_iadb/gwn2gbn_img10400_step175.png}
\end{tikzpicture}
&
\begin{tikzpicture}[scale=\scalevalue]
\PlotSingleImage{cat_res128_iadb/gwn2gbn_img10400_step225.png}
\end{tikzpicture}
&
\begin{tikzpicture}[scale=\scalevalue]
\PlotSingleImage{cat_res128_iadb/gwn2gbn_img10400_step250.png}
\end{tikzpicture}
\\[-0.4mm]
~ &
$t=250$ & 
$t=100$ &
$t=75$ &
$t=25$ &
$t=0$
\\[-0.4mm]
\end{tabular}
\end{tabular}
    \vspace{-2.5mm} 
    \caption{
        Image generation comparisons between DDIM, IADB and Ours trained on CelebA ($128^2$) and AFHQ-Cat ($128^2$) datasets, respectively.
        All methods start with the same initial Gaussian noise during the backward process.
        Our method generates more realistic content around the hair, eye, mouth regions compared to IADB. Compared to DDIM, our method achieves similar visual quality. 
        The impact of time-varying noise (we use $\tau=0.2$ in~\cref{eq:get_alpha}) can be seen by comparing IADB and ours starting from around $t=75$.
    }
    \label{fig:celeba_res128_sequence_gwn_gbn}
\end{figure*}

\end{document}



\title{Blue noise for diffusion models: Supplemental document}

\author{Xingchang Huang}
\affiliation{%
  \institution{MPI Informatics, VIA Center}
  \city{Saarbr{\"u}cken}
  \country{Germany}
}
\email{xhuang@mpi-inf.mpg.de}

\author{Corentin Sala\"un}
\affiliation{%
  \institution{MPI Informatics}
  \city{Saarbr{\"u}cken}
  \country{Germany}
}
\email{csalaun@mpi-inf.mpg.de}

\author{Cristina Vasconcelos}
\affiliation{%
  \institution{Google DeepMind}
  \city{London}
  \country{UK}
}
\email{crisnv@google.com}

\author{Christian Theobalt}
\affiliation{%
  \institution{MPI Informatics, VIA Center}
  \city{Saarbr{\"u}cken}
  \country{Germany}
}  
\email{theobalt@mpi-inf.mpg.de}

\author{Cengiz {\"O}ztireli}
\affiliation{%
  \institution{Google Research, University of Cambridge}
  \city{Cambridge}
  \country{UK}
}
\email{cengizo@google.com}

\author{Gurprit Singh}
\affiliation{%
  \institution{MPI Informatics, VIA Center}
  \city{Saarbr{\"u}cken}
  \country{Germany}
}  
\email{gsingh@mpi-inf.mpg.de}

\renewcommand{\shortauthors}{X. Huang, C. Sala\"un, C. Vasconcelos, C. Theobalt, C. {\"O}ztireli, G. Singh}











\begin{abstract}
In this document, we present additional details and results.
\end{abstract}




\maketitle


\newcommand{\GaussianNoiseSupp}{\bm{\epsilon}}

\section{Derivation of our backward process}

The following shows the derivation of Eq. (7) in the main paper. 
The goal is to compute $\Data_{t-1}$ using $\Data_{t}$ based on the definition of the backward process.
We start from:
\begin{align}
\Data_{t-1} &= \alphatminusone (\LMatrix_{t-1}\GaussianNoiseSupp) + (1 - \alphatminusone) \Data_0 
\end{align}
Next, we need to construct $\Data_{t}$ on the right hand side (RHS) as the following:
\begin{align}
    \Data_{t-1} &= \alphatminusone (\LMatrix_{t-1}\GaussianNoiseSupp) + (1 - \alphatminusone) \Data_0 
    \nonumber
    \\
    &= (\alphatminusone + \alphat - \alphat) (\LMatrix_{t-1}\GaussianNoiseSupp) + (1 - \alphatminusone + \alphat - \alphat) \Data_0
    \nonumber
    \\
    &= \alphat(\LMatrix_{t-1}\GaussianNoiseSupp) + (\alphatminusone - \alphat)(\LMatrix_{t-1}\GaussianNoiseSupp) + (1 - \alphat)\Data_0 + (\alphat - \alphatminusone)\Data_0
\end{align}
The $\LMatrix_{t-1}$ term can be expanded as:
\begin{align}
\label{eq:expandL}
\LMatrix_{t-1} &= \gamma_{t-1}\LMatrix_w + (1 - \gamma_{t-1})\LMatrix_b
\nonumber
\\
&= \LMatrix_b + \gamma_{t-1}(\LMatrix_w - \LMatrix_b)
\nonumber
\\
&= \LMatrix_b + (\gamma_{t-1} + \gamma_{t} - \gamma_{t})(\LMatrix_w - \LMatrix_b)
\nonumber
\\
&= \LMatrix_b + \gamma_{t}(\LMatrix_w - \LMatrix_b) + (\gamma_{t-1} - \gamma_{t})(\LMatrix_w - \LMatrix_b)
\end{align}
Based on above and $\Data_{t} = \alphat (\LMatrix_{t}\GaussianNoiseSupp) + (1 - \alphat) \Data_0$, we have:
\begin{align}
\Data_{t-1} &= \alphat(\LMatrix_t\GaussianNoiseSupp) + \alphat(\gamma_{t-1} - \gamma_{t})(\LMatrix_w - \LMatrix_b)\GaussianNoiseSupp + (\alphatminusone - \alphat)(\LMatrix_t\GaussianNoiseSupp)
\nonumber
\\
&+ (\alphatminusone - \alphat)(\gamma_{t-1} - \gamma_t)(\LMatrix_w - \LMatrix_b)\GaussianNoiseSupp + (1 - \alphat)\Data_0 + (\alphat - \alphatminusone)\Data_0
\nonumber
\\
&= \Data_t + (\alphat - \alphatminusone)(\Data_0 - \LMatrix_t\GaussianNoiseSupp) + (\gamma_{t-1} - \gamma_t)(\alphatminusone)(\LMatrix_w - \LMatrix_b)\GaussianNoiseSupp
\nonumber
\\
&= \Data_t + (\alphat - \alphatminusone)(\Data_0 - \LMatrix_t\GaussianNoiseSupp) + (\gamma_{t} - \gamma_{t-1})(\alphatminusone)(\LMatrix_b - \LMatrix_w)\GaussianNoiseSupp
\end{align}
We ensure that $\alphat \geq \alphatminusone$, $\gamma_{t} \geq \gamma_{t-1}$, $0 \leq \alphat \leq 1$, $0 \leq \gamma_t \leq 1$ for different schedulers.

The training procedure of our method is based on the derived backward process. For the terms $\LMatrix_t\GaussianNoiseSupp$, $(\LMatrix_w - \LMatrix_b)\GaussianNoiseSupp$,
when $\LMatrix_w$ represents the identity matrix, we do not need to actually perform the time-consuming matrix-vector multiplication for $\LMatrix_w\GaussianNoiseSupp$, as in this case $\GaussianNoiseSupp = \LMatrix_w\GaussianNoiseSupp$. Therefore, this only term that can introduce overhead is $\LMatrix_b\GaussianNoiseSupp$. We experimentally observed that this overhead is negligible for image resolutions at $64^2$, $128^2$, $256^2$.

\paragraph{Extension to DDIM}
\revision{
Here we show that our time-varying noise model can also extend to DDIM, following the procedure shown in~\citet{heitz2023iterative}.
First, we define:
\begin{align}
    y_{t} = (1 - \beta_{t}) x_0 + \beta_{t} (L_{t} \epsilon)
\end{align}
where $y_t, \beta_{t}$ are temporally introduced for easier derivations and will be replaced back by $x_t, \alpha_{t}$ later. Then, we have:
\begin{align}
\label{eq:ours_ddim_first}
    y_{t-1} &= (1 - \beta_{t-1}) x_0 + \beta_{t-1} (L_{t-1} \epsilon) \nonumber \\
    &= (1 - \beta_{t-1}) \frac{y_{t} - \beta_{t}(L_{t} \epsilon)}{1 - \beta_{t}} + \beta_{t-1} (L_{t-1} \epsilon) \nonumber \\
    &= \frac{1 - \beta_{t-1}}{1 - \beta_{t}} y_{t} - \frac{(1 - \beta_{t-1})\beta_{t}(L_t \epsilon)}{1 - \beta_{t}} + \beta_{t-1} (L_{t-1} \epsilon)
\end{align}
Based on~\cref{eq:expandL}, we can expand $\beta_{t-1}(L_{t-1}\epsilon)$ as two terms:
\begin{align}
    \beta_{t-1}(L_{t-1}\epsilon) &= \beta_{t-1}(L_b + \gamma_{t}(L_w - L_b) + (\gamma_{t-1}-\gamma_{t})(L_w - L_b)\epsilon \nonumber \\
    &= \beta_{t-1}(L_{t} \epsilon) + \beta_{t-1}(\gamma_{t-1} - \gamma_{t}) (L_w - L_b) \epsilon
\end{align}
By merging the second and third terms on the right-hand side (RHS) for the expanded version of~\cref{eq:ours_ddim_first}, we can get:
\begin{align}
    y_{t-1} &= \frac{1 - \beta_{t-1}}{1 - \beta_{t}} y_{t} - \frac{L_{t} \epsilon (\beta_{t} - \beta_{t-1})}{1 - \beta_{t}} + \beta_{t-1}(\gamma_{t-1}-\gamma_{t})(L_w - L_b)\epsilon \nonumber \\
    &= \frac{1 - \beta_{t-1}}{1 - \beta_{t}} (y_{t} - L_{t} \epsilon) + L_{t} \epsilon + \beta_{t-1}(\gamma_{t-1} - \gamma_{t})(L_w - L_b) \epsilon
\end{align}
Next, we let $\beta_{t} = \frac{\sqrt{1 - \overline{\alpha}_{t}}}{\sqrt{\overline{\alpha}_{t}} + \sqrt{1 - \overline{\alpha}_{t}}}$
and 
$y_{t} = \frac{x_{t}}{\sqrt{\overline{\alpha}_{t}} + \sqrt{1 - \overline{\alpha}_{t}}}$
where $\overline{\alpha}_{t} = \prod_{s=1}^{t}\alpha_{s}$.
Lastly, we can derive the backward process as the following:
\begin{align}
    x_{t-1} &= \frac{\sqrt{\overline{\alpha}_{t-1}}}{\sqrt{\overline{\alpha}_{t}}} (x_{t} - \frac{\sqrt{\overline{\alpha}_{t}}\sqrt{1 - \overline{\alpha}_{t-1}} - \sqrt{\overline{\alpha}_{t-1}}\sqrt{1 - \overline{\alpha}_{t}}}{\sqrt{\overline{\alpha}_{t-1}}} L_{t}\epsilon) \nonumber \\ 
    &+ (\gamma_{t} - \gamma_{t-1})\sqrt{1 - \overline{\alpha}_{t-1}}(L_b - L_w)\epsilon
\end{align}
In this case, the network needs to learn $L_{t} \epsilon$ and $\sqrt{1 - \overline{\alpha}_{t-1}}(L_b - L_w)\epsilon$.
}

\section{Datasets, network architecture and training details}

\revision{
We use the following datasets for unconditional image generation: CelebA ($64^2$ and $128^2$ resolutions, 30,000 training images)~\cite{CelebAMask-HQ}, AFHQ-Cat ($64^2$ and $128^2$ resolutions, 5,153 training images)~\cite{choi2020stargan} and LSUN-Church ($64^2$ resolution, 30,000 out of 126,227 training images)~\cite{yu15lsun}. These partitions were set in order to replicate the training conditions of the methods compared here. For conditional image generation, we conduct experiments on image super-resolution using CelebA from $64^2$ to $128^2$, $32^2$ to $128^2$ and LSUN-Church from $32^2$ to $128^2$. For both datasets, we use 25,000 images for training and 5,000 images for evaluation.
}

We use the diffuser library~\cite{von-platen-etal-2022-diffusers} to build the 2D U-Net~\cite{ronneberger2015u} architecture with 6, 7 down- and up-sampling layers with skip connections for image resolution of $64^2$, $128^2$, respectively. 
The number of channels we use are (128, 128, 256, 256, 512, 512) for 6 layers and (128, 128, 128, 256, 256, 512, 512) for 7 layers.
Self-attention~\cite{vaswani2017attention} module is added in the second last of the down-sampling layer and the second of the up-sampling layer.

For network training details, we show the number of epochs and batch size used for training on different datasets for DDIM, IADB and Ours in~\cref{tab:training}.
~\Cref{tab:training} also includes the $\tau$ (Eq. (9) in the main paper) values we use for all experiments.
For the CelebA ($64^2$) experiment, we use the exact linear scheduler for $\gamma_t$.

\begin{table}[h]
\centering
\caption{
Network training details and the choices of $\tau$ (Eq. (9) in the main paper) for our method.
For the CelebA ($64^2$) experiment, we use the exact linear scheduler for $\gamma_t$.
\revision{
We use latent diffusion model (LDM)~\cite{rombach2022high} for the AFHQ-Cat ($512^2$) experiment.
}
}
\begin{tabular}{cccc}
\toprule
Dataset & \#Epochs & Batchsize & $\tau$ \\
\midrule
AFHQ-Cat ($64^2$) & 1,000 & 256 & 1,000 \\ 
AFHQ-Cat ($128^2$) & 1,000 & 128 & 0.2\\ 
\revision{AFHQ-Cat (LDM, $512^2$)} & 1,000 & 256 & 1,000 \\
CelebA ($64^2$) & 1,000 & 256 & (linear) \\ 
CelebA ($128^2$) & 700 & 128 & 0.2 \\ 
LSUN-Church ($64^2$) & 1,000 & 256 & 1,000 \\ 
CelebA ($64^2\rightarrow128^2$) & 400 & 128 & 0.2 \\ 
CelebA ($32^2\rightarrow128^2$) & 80 & 64 & 0.2 \\
LSUN-Church ($32^2\rightarrow128^2$) & 80 & 64 & 0.2 \\
\bottomrule
\end{tabular}
\label{tab:training}
\end{table}
\begin{table}[t!]
\centering
\smaller
\smaller
\caption{
Early stopping tests using Gaussian noise only and Gaussian blue noise only on AFHQ-Cat($128^2$). $T_e$ represents the time step we apply early stopping. When $T_e=0$, the model falls back to the full backward process.
}
\begin{tabular}{c|ccc|ccc}
\toprule
& \multicolumn{3}{c|}{Ours (Gaussian noise only)} & \multicolumn{3}{c}{Ours (Gaussian blue noise only)} \\
$T_e$ & FID ($\downarrow$) & Precision ($\uparrow$) & Recall ($\uparrow$) & FID ($\downarrow$) & Precision ($\uparrow$) & Recall ($\uparrow$) \\
\hline
200 & 24.10 & 0.26 & 0.08 & \textbf{20.31} & \textbf{0.41} & \textbf{0.11} \\
150 & \textbf{17.39} & \textbf{0.47} & \textbf{0.13} & 22.42 & 0.33 & 0.12 \\
0 & \textbf{10.81} & \textbf{0.78} & \textbf{0.31} & 17.61 & 0.59 & 0.18 \\
\bottomrule
\end{tabular}
\label{tab:earlystopping}
\end{table}

\begin{table}[t!]
\centering
\smaller
\caption{
Image super-resolution metrics using IADB and Ours. Our method shows consistent improvement over IADB according to the SSIM and PSNR metrics. 2x and 4x represent super-resolution from $64^2$ to $128^2$ and $32^2$ to $128^2$, respectively.
}
\begin{tabular}{c|cc|cc}
\toprule
& \multicolumn{2}{c|}{IADB} & \multicolumn{2}{c}{Ours} \\
 & SSIM ($\uparrow$) & PSNR ($\uparrow$) & SSIM ($\uparrow$) & PSNR ($\uparrow$) \\
\hline
CelebA ($2\times$) & 0.91 & 30.73 & \textbf{0.92} & \textbf{31.56} \\
CelebA ($4\times$) & 0.76 &  24.74 &  \textbf{0.77} & \textbf{25.03} \\
LSUN-Church ($4\times$) & 0.57 & 19.46 & \textbf{0.59} & \textbf{20.00} \\
\bottomrule
\end{tabular}
\label{tab:superres}
\end{table}

\section{Additional results}

\paragraph{Gaussian blue noise at different resolutions.}
\Cref{fig:gbn_different_resolutions} visualizes Gaussian blue noise masks at different resolution using our padding strategy. 
For the masks at resolution $128^2$ and $256^2$, the seams between the padded $64^2$ tiles are not visible and the property of blue noise is still preserved.
\revision{In addition, we show in~\cref{fig:gbn_repetitive} that padding/tiling with the same $64^2$ Gaussian blue noise mask results in repetitive noise patterns, as well as structural artifacts in the frequency spectra.}

\begin{figure}[h!]

    \centering

\newcommand{\PlotSingleImage}[1]{%
        \begin{scope}
            \clip (0,0) -- (2.5,0) -- (2.5,2.5) -- (0,2.5) -- cycle;
            \path[fill overzoom image=figures/noise_visualization/#1] (0,0) rectangle (2.5cm,2.5cm);
        \end{scope}
        \draw (0,0) -- (2.5,0) -- (2.5,2.5) -- (0,2.5) -- cycle;
}

\newcommand{\PlotImageAndCrop}[2]{%
    \begin{scope}
        \clip (0,0) -- (3.2,0) -- (3.2,3.25)-- (0.0,3.25) -- cycle;
        \path[fill overzoom image=figures/noise_visualization/#1] (0,0) rectangle (3.25,3.25);
    \end{scope}
    \begin{scope}
        \clip (2.0,0.1) -- (3.1,0.1) -- (3.1,1.2) -- (2.0,1.2) -- cycle;
        \path[fill overzoom image=figures/noise_visualization/#2] (2.0,0.1) rectangle (3.1cm,1.2cm);
    \end{scope}
}

\newcommand{\TwoColumnFigure}[2]{%
    \begin{tabular}{c@{\;}c@{}}
        \hspace*{-2.5mm}
        \begin{tikzpicture}[scale=0.563]
            \PlotSingleImage{#1}
        \end{tikzpicture}
         & 
         \begin{tikzpicture}[scale=0.563]
            \PlotSingleImage{#2}
        \end{tikzpicture}
    \end{tabular}%
}

\newcommand{\scaleval}{0.89} 

\small
\hspace*{-4mm}
\begin{tabular}{c@{\;}c@{\;}c@{\;}c@{}}
~ &
Gaussian blue noise: $64^2$ &
$128^2$ &
$256^2$
\\
&
\begin{tikzpicture}[scale=\scaleval]
    \PlotImageAndCrop{bn_res64.png}{bn_res64_ps.png}
\end{tikzpicture}
&
\begin{tikzpicture}[scale=\scaleval]
    \PlotImageAndCrop{bn_res128.png}{bn_res128_ps.png}
\end{tikzpicture}
&
\begin{tikzpicture}[scale=\scaleval]
    \PlotImageAndCrop{bn_res256.png}{bn_res256_ps.png}
\end{tikzpicture}
\end{tabular} 
    \caption{
        Gaussian blue noise masks at different resolutions using our padding strategy and the corresponding frequency power spectra. Padding multiple $(64^2)$ tiles does not produce visible artifact and have unnoticeable impact on the frequency distribution.
    }
    \label{fig:gbn_different_resolutions}
\end{figure}

\begin{figure}[h!]
    \centering

\newcommand{\PlotSingleImage}[1]{%
        \begin{scope}
            \clip (0,0) -- (2.5,0) -- (2.5,2.5) -- (0,2.5) -- cycle;
            \path[fill overzoom image=figures/more_ablations/#1] (0,0) rectangle (2.5cm,2.5cm);
        \end{scope}
        \draw (0,0) -- (2.5,0) -- (2.5,2.5) -- (0,2.5) -- cycle;
}

\newcommand{\PlotImageAndCrop}[2]{%
    \begin{scope}
        \clip (0,0) -- (3.2,0) -- (3.2,3.25)-- (0.0,3.25) -- cycle;
        \path[fill overzoom image=figures/more_ablations/#1] (0,0) rectangle (3.25,3.25);
    \end{scope}
    \begin{scope}
        \clip (2.0,0.1) -- (3.1,0.1) -- (3.1,1.2) -- (2.0,1.2) -- cycle;
        \path[fill overzoom image=figures/more_ablations/#2] (2.0,0.1) rectangle (3.1cm,1.2cm);
    \end{scope}
}

\newcommand{\TwoColumnFigure}[2]{%
    \begin{tabular}{c@{\;}c@{}}
        \hspace*{-2.5mm}
        \begin{tikzpicture}[scale=0.563]
            \PlotSingleImage{#1}
        \end{tikzpicture}
         & 
         \begin{tikzpicture}[scale=0.563]
            \PlotSingleImage{#2}
        \end{tikzpicture}
    \end{tabular}%
}

\newcommand{\scaleval}{1.25} 

\small
\hspace*{-4mm}
\begin{tabular}{c@{\;}c@{\;}c@{}}
~ &
$128^2$ &
$256^2$
\\
&
\begin{tikzpicture}[scale=\scaleval]
    \PlotImageAndCrop{gaussianBN_res128_and_spectrum_0_repetitive_True_noise.png}{gaussianBN_res128_and_spectrum_0_repetitive_True_spectrum.png}
\end{tikzpicture}
&
\begin{tikzpicture}[scale=\scaleval]
    \PlotImageAndCrop{gaussianBN_res256_and_spectrum_0_repetitive_True_noise.png}{gaussianBN_res256_and_spectrum_0_repetitive_True_spectrum.png}
\end{tikzpicture}
\end{tabular} 
    \caption{
        \revision{Padding/tiling with the same $64^2$ Gaussian blue noise mask to generate higher-resolution masks results in repetitive noise patterns, as well as structural artifacts in the frequency spectra.}
    }
    \label{fig:gbn_repetitive}
\end{figure}

\paragraph{Early stopping tests.}
Though using Gaussian blue noise only leads to worse results in terms of both quantitative and qualitative evaluations, we observe that the image content appears more clear visually in the early time steps.
Based on this observation, we perform a study on early stopping where we stop at a specific early time step $T_e$ and reconstruct the final image with one step.
As shown in~\cref{tab:earlystopping}, early stopping at $T_e=200$ gives much better results using Gaussian blue noise only.
However, the quality does not improve with more steps but fluctuates in terms of the metrics.

\paragraph{Nearest neighbors test.}
We conduct a nearest neighbors test on AFHQ-Cat ($64^2$) using our method to check if there exists an over-fitting issue.
We test on AFHQ-Cat ($64^2$) as it is the dataset with smallest training samples and lowest resolution among our tested datasets. 
As shown in~\cref{fig:cat_res64_nearest_neighbors}, though the nearest neighbors (second to fifth columns) may be semantically similar to the query on the leftmost column, the generated samples are not identical to the training set samples.
This means our method does not suffer overfit the training data.

\paragraph{Image super-resolution.}
\Cref{tab:superres} provides the SSIM and PSNR metrics for the image super-resolution tasks using IADB and Ours.
Our method consistenly improves over IADB.

\paragraph{Image generation.}
We provide the full~\cref{tab:quantitative} for quantitative comparisons including FID, Precision and Recall on image generation tasks. More results, comparisons and interactive visualization can be found in the Supplemental HTML.

\paragraph{Extension to DDIM~\cite{song2020denoising}}
\revision{
Detailed derivations of extending our time-varying noise model to DDIM can be found in Supplemental document Sec. 1.
\cref{tab:extension_to_ddim} shows preliminary results on AHFQ-Cat ($64^2$) and Ours (DDIM) gets better FID than the original DDIM.
}

\paragraph{Extension to LDM~\cite{rombach2022high}.}
\revision{
Our time-varing noise model can also be incorporated into latent diffusion model (LDM)~\cite{rombach2022high} for high-resolution image generation.
We compare IADB and Ours used for latent diffusion in~\cref{tab:extension_to_ldm} and the preliminary results show that Ours gets better FID than IADB.
The visual comparisons can be found in the main paper.
}

\paragraph{Ablation on $\gamma$-scheduler.}
\revision{
\Cref{tab:ablation_gamma_scheduler} compares different parameters and functions of our $\gamma$-scheduler. For our sigmoid-based $\gamma$-scheduler, $\gamma=0.02$ gives better results than $\gamma=1000$, showing the importance of choosing the $\gamma$ value for our $\gamma$-scheduler. Using the cosine-based function proposed by~\cite{nichol2021improved} resulted in worse FID, showing the importance of choosing the function for our $\gamma$-scheduler.
}

\begin{figure}
\centering
\includegraphics[width=1.0\columnwidth]{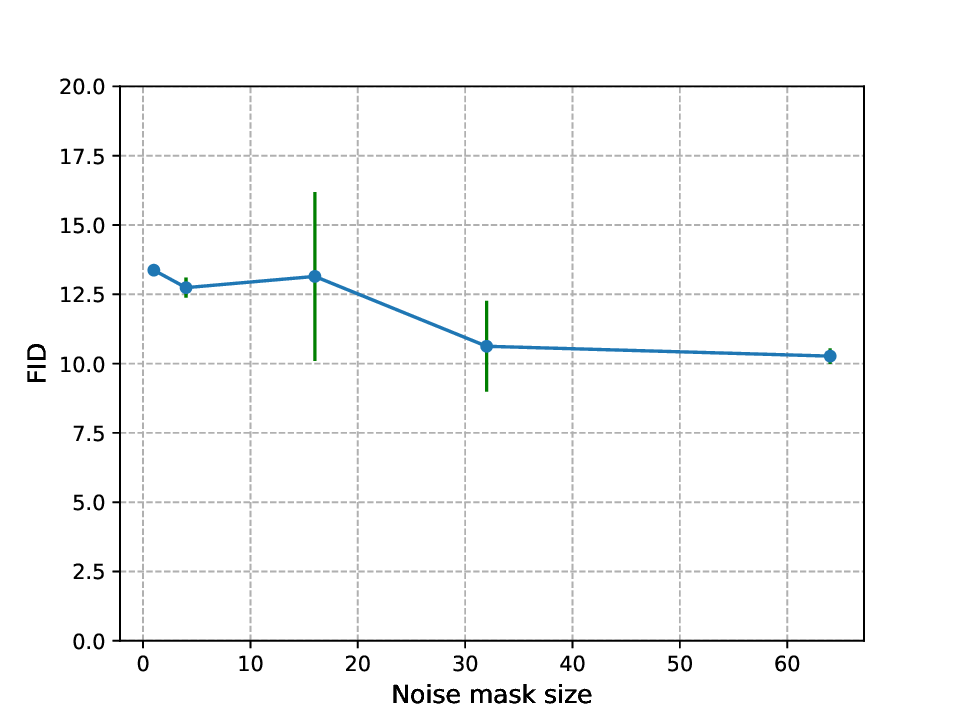}
\caption{
\revision{
Ablation study on different Gaussian blue noise sizes used for padding/tiling ($1^2$, $4^2$, $16^2$, $32^2$, $64^2$) experimented on the AFHQ-Cat ($128^2$) dataset. Note that $1^2$ size is equivalent to use Gaussian (white) noise only. We use $64^2$ size as our method in all experiments.
}
}
\label{fig:ablation_noise_mask_size}
\end{figure}

\paragraph{Ablation on noise mask size for padding/tiling}
\revision{
\Cref{fig:ablation_noise_mask_size} shows an ablation study on padding/tiling using different Gaussian blue noise sizes ($1^2$, $4^2$, $16^2$, $32^2$, $64^2$) on the AFHQ-Cat ($128^2$) dataset. Note that $1^2$ size is equivalent to use Gaussian (white) noise. We use $64^2$ size as our method for all experiments.
The figure shows that increasing the size of Gaussian blue noise mask leads to lower FID than using Gaussian (white) noise in terms of mean values of multiple experiments. However, the standard deviations of using tiled Gaussian blue noise mask can be higher than using Gaussian (white) noise.
}

\paragraph{Timing}
\revision{
Here, we present the timing results obtained using a single RTX 2080 NVIDIA GPU for our pipeline. To assess the average inference time for both IADB and our networks, we conducted tests with a batch size of 1 and $\TotalTimesteps= 250$. The network architectures are identical, except for our network having a 6-channel output instead of 3. The average network inference time is approximately 0.020 seconds for generating a $64^2$ image and around 0.023 seconds for a $128^2$ image, applicable to both IADB and our method. Regarding noise generation timing, our approach takes roughly 0.0001 seconds to generate a Gaussian blue noise mask at a resolution of $64^2$ and about 0.0002 seconds to generate a Gaussian noise or Gaussian blue noise at a resolution of $128^2$.
}

\begin{table}[t]
\centering
\caption{
\revision{
Preliminary results on comparing DDIM~\cite{song2020denoising} and Ours (DDIM) on AFHQ-Cat ($64^2$).
}
}
\resizebox{5.4cm}{!}{
\begin{tabular}{c|>{\centering\arraybackslash}m{0.1\textwidth}>{\centering\arraybackslash}m{0.1\textwidth}>
{\centering\arraybackslash}m{0.1\textwidth}}
\toprule
Method & DDIM & Ours (DDIM) \\
\midrule
FID ($\downarrow$) & 9.82 & \textbf{7.11} \\
\bottomrule
\end{tabular}
}
\label{tab:extension_to_ddim}
\end{table}

\begin{table}[t]
\centering
\caption{
\revision{
Preliminary results on comparing IADB and Ours in latent diffusion model (LDM)~\cite{rombach2022high} on AFHQ-Cat ($512^2$).
}
}
\resizebox{6.0cm}{!}{
\begin{tabular}{c|>{\centering\arraybackslash}m{0.1\textwidth}>{\centering\arraybackslash}m{0.1\textwidth}>
{\centering\arraybackslash}m{0.1\textwidth}}
\toprule
Method for LDM & IADB & Ours \\
\midrule
FID ($\downarrow$) & 12.19 & \textbf{11.45} \\
\bottomrule
\end{tabular}
}
\label{tab:extension_to_ldm}
\end{table}

\begin{table}[t]
\centering
\caption{
\revision{
Comparing the impact of $\gamma$-scheduler on CelebA ($128^2$). 
Our $\gamma$-scheduler with $\gamma=0.2$ results in lower FID than $\gamma=1000$ and cosine-based scheduler~\cite{nichol2021improved}.
}
}
\resizebox{8.4cm}{!}{
\begin{tabular}{c|>{\centering\arraybackslash}m{0.1\textwidth}>{\centering\arraybackslash}m{0.1\textwidth}>{\centering\arraybackslash}m{0.1\textwidth}>
{\centering\arraybackslash}m{0.1\textwidth}}
\toprule
$\gamma$-scheduler & cosine-based & $\gamma=1000$ & $\gamma=0.2$ \\
\midrule
FID ($\downarrow$) & 37.13 & 29.70 & \textbf{16.38} \\
\bottomrule
\end{tabular}
}
\label{tab:ablation_gamma_scheduler}
\end{table}

\begin{figure}[h]
    \centering

\newcommand{\PlotSingleImage}[1]{%
        \begin{scope}
            \clip (0,0) -- (2.5,0) -- (2.5,2.5) -- (0,2.5) -- cycle;
            \path[fill overzoom image=figures/cat_res128_earlystopping/#1] (0,0) rectangle (2.5cm,2.5cm);
        \end{scope}
        \draw (0,0) -- (2.5,0) -- (2.5,2.5) -- (0,2.5) -- cycle;
}

\newcommand{\TwoColumnFigure}[2]{%
    \begin{tabular}{c@{\;}c@{}}
        \hspace*{-2.5mm}
        \begin{tikzpicture}[scale=0.563]
            \PlotSingleImage{#1}
        \end{tikzpicture}
         & 
         \begin{tikzpicture}[scale=0.563]
            \PlotSingleImage{#2}
        \end{tikzpicture}
    \end{tabular}%
}

\newcommand{\scaleval}{0.64}    
\small
\hspace*{-2.5mm}  
\begin{tabular}{c@{\;}c@{\;}c@{\;}c@{\;}c@{\;}c@{}}
~ &
\\
\rotatebox{90}{\hspace{0.15cm} \scriptsize Random noise}
&
\begin{tikzpicture}[scale=\scaleval]
    \PlotSingleImage{gwn_earlystop200_img00000_step75.png}
\end{tikzpicture}
&
\begin{tikzpicture}[scale=\scaleval]
    \PlotSingleImage{gwn_earlystop200_img01000_step75.png}
\end{tikzpicture}
&
\begin{tikzpicture}[scale=\scaleval]
    \PlotSingleImage{gwn_earlystop200_img01200_step75.png}
\end{tikzpicture}
&
\begin{tikzpicture}[scale=\scaleval]
    \PlotSingleImage{gwn_earlystop200_img02400_step75.png}
\end{tikzpicture}
&
\begin{tikzpicture}[scale=\scaleval]
    \PlotSingleImage{gwn_earlystop200_img03000_step75.png}
\end{tikzpicture}
\\[-0.4mm]
\rotatebox{90}{\hspace{0.25cm} \scriptsize Blue noise}
&
\begin{tikzpicture}[scale=\scaleval]
    \PlotSingleImage{gbn_earlystop200_img00000_step75.png}
\end{tikzpicture}
&
\begin{tikzpicture}[scale=\scaleval]
    \PlotSingleImage{gbn_earlystop200_img01000_step75.png}
\end{tikzpicture}
&
\begin{tikzpicture}[scale=\scaleval]
    \PlotSingleImage{gbn_earlystop200_img01200_step75.png}
\end{tikzpicture}
&
\begin{tikzpicture}[scale=\scaleval]
    \PlotSingleImage{gbn_earlystop200_img03000_step75.png}
\end{tikzpicture}
&
\begin{tikzpicture}[scale=\scaleval]
    \PlotSingleImage{gbn_earlystop200_img02400_step75.png}
\end{tikzpicture}
\\[-0.4mm]
\end{tabular} 
    \caption{
        Early stopping test on AFHQ-Cat ($128^2$) using single noise during the training process. The backward process is stopped at $t=200$ step. 
        First row shows generated results using Gaussian noise and the second using Gaussian blue noise. While blue noise alone creates some low frequency artifacts in the eye region, it generates sharper details than random noise.
    }
    \label{fig:cat_res128_earlystopping}
\end{figure}

\begin{figure}[t!]
    \centering

\newcommand{\PlotSingleImage}[1]{%
        \begin{scope}
            \clip (0,0) -- (2.5,0) -- (2.5,2.5) -- (0,2.5) -- cycle;
            \path[fill overzoom image=figures/cat_res64_nearest_neighbors/#1] (0,0) rectangle (2.5cm,2.5cm);
        \end{scope}
        \draw (0,0) -- (2.5,0) -- (2.5,2.5) -- (0,2.5) -- cycle;
}

\newcommand{\TwoColumnFigure}[2]{%
    \begin{tabular}{c@{\;}c@{}}
        \hspace*{-2.5mm}
        \begin{tikzpicture}[scale=0.563]
            \PlotSingleImage{#1}
        \end{tikzpicture}
         & 
         \begin{tikzpicture}[scale=0.563]
            \PlotSingleImage{#2}
        \end{tikzpicture}
    \end{tabular}%
}

\newcommand{\scaleval}{0.655}    
\small
\hspace*{-4mm}
\begin{tabular}{c@{\;}c@{\;}c@{\;}c@{\;}c@{\;}c@{}}
 ~ &
 Ours &
 1st-nearest &
 2nd-nearest &
 3rd-nearest &
 4th-nearest
\\
&
\begin{tikzpicture}[scale=\scaleval]
    \PlotSingleImage{cat_res64_img00001.png}
\end{tikzpicture}
&
\begin{tikzpicture}[scale=\scaleval]
    \PlotSingleImage{cat_res64_img00001_nn0.png}
\end{tikzpicture}
&
\begin{tikzpicture}[scale=\scaleval]
    \PlotSingleImage{cat_res64_img00001_nn1.png}
\end{tikzpicture}
&
\begin{tikzpicture}[scale=\scaleval]
    \PlotSingleImage{cat_res64_img00001_nn2.png}
\end{tikzpicture}
&
\begin{tikzpicture}[scale=\scaleval]
    \PlotSingleImage{cat_res64_img00001_nn3.png}
\end{tikzpicture}
\\[-0.4mm]
&
\begin{tikzpicture}[scale=\scaleval]
    \PlotSingleImage{cat_res64_img00003.png}
\end{tikzpicture}
&
\begin{tikzpicture}[scale=\scaleval]
    \PlotSingleImage{cat_res64_img00003_nn0.png}
\end{tikzpicture}
&
\begin{tikzpicture}[scale=\scaleval]
    \PlotSingleImage{cat_res64_img00003_nn1.png}
\end{tikzpicture}
&
\begin{tikzpicture}[scale=\scaleval]
    \PlotSingleImage{cat_res64_img00003_nn2.png}
\end{tikzpicture}
&
\begin{tikzpicture}[scale=\scaleval]
    \PlotSingleImage{cat_res64_img00003_nn3.png}
\end{tikzpicture}
\\[-0.4mm]
&
\begin{tikzpicture}[scale=\scaleval]
    \PlotSingleImage{cat_res64_img00004.png}
\end{tikzpicture}
&
\begin{tikzpicture}[scale=\scaleval]
    \PlotSingleImage{cat_res64_img00004_nn0.png}
\end{tikzpicture}
&
\begin{tikzpicture}[scale=\scaleval]
    \PlotSingleImage{cat_res64_img00004_nn1.png}
\end{tikzpicture}
&
\begin{tikzpicture}[scale=\scaleval]
    \PlotSingleImage{cat_res64_img00004_nn2.png}
\end{tikzpicture}
&
\begin{tikzpicture}[scale=\scaleval]
    \PlotSingleImage{cat_res64_img00004_nn3.png}
\end{tikzpicture}
\\[-0.4mm]
&
\begin{tikzpicture}[scale=\scaleval]
    \PlotSingleImage{cat_res64_img00005.png}
\end{tikzpicture}
&
\begin{tikzpicture}[scale=\scaleval]
    \PlotSingleImage{cat_res64_img00005_nn0.png}
\end{tikzpicture}
&
\begin{tikzpicture}[scale=\scaleval]
    \PlotSingleImage{cat_res64_img00005_nn1.png}
\end{tikzpicture}
&
\begin{tikzpicture}[scale=\scaleval]
    \PlotSingleImage{cat_res64_img00005_nn2.png}
\end{tikzpicture}
&
\begin{tikzpicture}[scale=\scaleval]
    \PlotSingleImage{cat_res64_img00005_nn3.png}
\end{tikzpicture}
\\[-0.4mm]
&
\begin{tikzpicture}[scale=\scaleval]
    \PlotSingleImage{cat_res64_img00007.png}
\end{tikzpicture}
&
\begin{tikzpicture}[scale=\scaleval]
    \PlotSingleImage{cat_res64_img00007_nn0.png}
\end{tikzpicture}
&
\begin{tikzpicture}[scale=\scaleval]
    \PlotSingleImage{cat_res64_img00007_nn1.png}
\end{tikzpicture}
&
\begin{tikzpicture}[scale=\scaleval]
    \PlotSingleImage{cat_res64_img00007_nn2.png}
\end{tikzpicture}
&
\begin{tikzpicture}[scale=\scaleval]
    \PlotSingleImage{cat_res64_img00007_nn3.png}
\end{tikzpicture}
\\[-0.4mm]
\end{tabular}
    \caption{
        We conduct a nearest neighbors test showing our method does not overfit the training data. 
        Generated samples of our method trained on AFHQ-Cat ($64^2$) are shown in the leftmost column. 
        Training set nearest neighbors are in the remaining columns, ordered from the 1st-nearest neighbor to the 4th-nearest neighbor (based on pixel-wise mean squared distance) from left to right.
    }
    \label{fig:cat_res64_nearest_neighbors}
\end{figure}


\begin{table*}[ht]
\centering
\caption{
Quantitative comparisons between DDIM~\cite{song2020denoising}, IADB~\cite{heitz2023iterative} and Ours on different datasets, \revision{including FID, Precision and Recall metrics.}
Our method shows consistent improvements over IADB in terms of FID score. Best score for each metric and dataset is shown in bold, second best is underlined.
}
\begin{tabular}{c|ccc|ccc|ccc}
\toprule
& \multicolumn{3}{c|}{DDIM} & \multicolumn{3}{c|}{IADB} & \multicolumn{3}{c}{Ours} \\
& FID ($\downarrow$) & Precision ($\uparrow$) & Recall ($\uparrow$) & FID ($\downarrow$) & Precision ($\uparrow$) & Recall ($\uparrow$) & FID ($\downarrow$) & Precision ($\uparrow$) & Recall ($\uparrow$) \\
\hline
AFHQ-Cat ($64^2$) & 9.82 & \textbf{0.83} & 0.32 & \underline{9.18} & 0.72 & \underline{0.37} & \textbf{7.95} & \underline{0.73} & \textbf{0.50} \\
AFHQ-Cat ($128^2$) & \underline{10.73} & \textbf{0.84} & 0.28 & 10.81 & \underline{0.78} & \underline{0.31} & \textbf{9.47} & \underline{0.78} & \textbf{0.34} \\
CelebA ($64^2$)  & 9.26 & 0.83 & 0.46 & \underline{7.53} & \underline{0.84} & \textbf{0.53} & \textbf{7.05} & \textbf{0.85} & \underline{0.52} \\
CelebA ($128^2$)  & \textbf{11.92} & \textbf{0.81} & \textbf{0.48} & 20.71 & 0.75 & \underline{0.45} & \underline{16.38} & \textbf{0.81} & 0.42 \\
LSUN-Church ($64^2$) & 16.46 & \underline{0.74} & 0.41 & \underline{13.12} & 0.71 & \underline{0.47} & \textbf{10.16} & \textbf{0.76} & \textbf{0.48} \\
\bottomrule
\end{tabular}
\label{tab:quantitative}
\end{table*}




\bibliographystyle{ACM-Reference-Format}
\bibliography{paper}